\documentclass[10pt,twocolumn,letterpaper]{article}

\usepackage{cvpr}              
\usepackage{amsmath,amssymb,amsfonts}
\usepackage{algorithmic}
\usepackage{graphicx}
\usepackage{textcomp}
\usepackage{xcolor}
\usepackage[table]{xcolor}
\usepackage{color,colortbl}
\usepackage{wrapfig}
\usepackage{array}
\usepackage{bbding}
\usepackage{arydshln}
\usepackage{amsmath}
\usepackage{amssymb}
\usepackage{graphicx}
\usepackage{booktabs}
\usepackage{CJKutf8}
\usepackage{multirow}
\usepackage{enumitem} 
\usepackage{makecell}
\usepackage{bbm}
\usepackage{arydshln}
\usepackage[accsupp]{axessibility}
\usepackage[normalem]{ulem}
\newcommand{\tabincell}[2]{\begin{tabular}{@{}#1@{}}#2\end{tabular}}

\newcommand{\tocite}[1]{{\textcolor{red}{TOCITE}}}

\newcolumntype{I}{!{\vrule width 1.2pt}}
\newlength\savedwidth
\newcommand\whline{\noalign{\global\savedwidth\arrayrulewidth
                            \global\arrayrulewidth 1.5pt}
                   \hline
                   \noalign{\global\arrayrulewidth\savedwidth}}
\newlength\savewidth

\newcommand{\s}[1]{}
\definecolor{darkergreen}{RGB}{21, 152, 56}
\newcommand\greenp[1]{\textcolor{darkergreen}{}}
\usepackage[linesnumbered,ruled,vlined]{algorithm2e}
\definecolor{cvprblue}{rgb}{0.21,0.49,0.74}
\definecolor{newcolor}{rgb}{.8,.349,.1}
\definecolor{mypink}{rgb}{.99,.91,.95}
\definecolor{mygreen}{RGB}{26,104,64}
\definecolor{lgreen}{HTML}{00B050}
\definecolor{mygray}{HTML}{eeeeee}
\definecolor{myred}{HTML}{fae4df}
\definecolor{mydarkred}{HTML}{ed3333}
\definecolor{mydarkgray}{HTML}{808080}
\definecolor{myblue}{HTML}{E0F0FA}
\definecolor{mypink}{rgb}{0,0,0}

\usepackage[pagebackref,
            breaklinks,
            colorlinks,
            allcolors=cvprblue,
            linkcolor=cvprblue,  
            anchorcolor=cvprblue,
            citecolor=cvprblue,
            ]{hyperref}

\def\paperID{*****} 
\def\confName{CVPR}
\def\confYear{2026}

\title{Prefill-Time Intervention for Mitigating Hallucination \\ in Large Vision-Language Models}

\author{
    Chengsheng Zhang\textsuperscript{1},
    Chenghao Sun\textsuperscript{1},
    Xinyan Jiang\textsuperscript{2,3},
    Wei Li\textsuperscript{1},
    Xinmei Tian\textsuperscript{1}\thanks{Corresponding author.},
    \\
    \textsuperscript{1}University of Science and Technology of China \\
    \textsuperscript{2}Shanghai Advanced Research Institute, Chinese Academy of Sciences, Shanghai, China \\
    \textsuperscript{3}University of Chinese Academy of Sciences, Beijing, China \\
    {\tt\small \{zhangcs66, chsun, lwzkd\}@mail.ustc.edu.cn}, 
    {\tt\small xinmei@ustc.edu.cn}
}

\begin{document}
\maketitle
\begin{abstract}
Large Vision-Language Models (LVLMs) have achieved remarkable progress in visual-textual understanding, yet their reliability is critically undermined by hallucinations, i.e., the generation of factually incorrect or inconsistent responses.
While recent studies using steering vectors demonstrated promise in reducing hallucinations, a notable challenge remains: they inadvertently amplify the severity of residual hallucinations. 
We attribute this to their exclusive focus on the decoding stage, where errors accumulate autoregressively and progressively worsen subsequent hallucinatory outputs.
To address this, we propose \textbf{P}refill-\textbf{T}ime \textbf{I}ntervention (\textbf{PTI}), a novel steering paradigm that intervenes only once during the prefill stage, enhancing the initial Key-Value (KV) cache before error accumulation occurs.
Specifically, PTI is modality-aware, deriving distinct directions for visual and textual representations. 
This intervention is decoupled to steer keys toward visually-grounded objects and values to filter background noise, correcting hallucination-prone representations at their source.
Extensive experiments demonstrate PTI's significant performance in mitigating hallucinations and its generalizability across diverse decoding strategies, LVLMs, and benchmarks. 
Moreover, PTI is orthogonal to existing decoding-stage methods, enabling plug-and-play integration and further boosting performance. 
Code is available at: \href{https://github.com/huaiyi66/PTI}{https://github.com/huaiyi66/PTI}.

\end{abstract}
    
\section{Introduction}
\label{sec:intro}

\begin{figure}[!ht] 
    \centering 
    \begin{minipage}[t]{\linewidth} 
        \centering
        \includegraphics[width=0.95\linewidth]{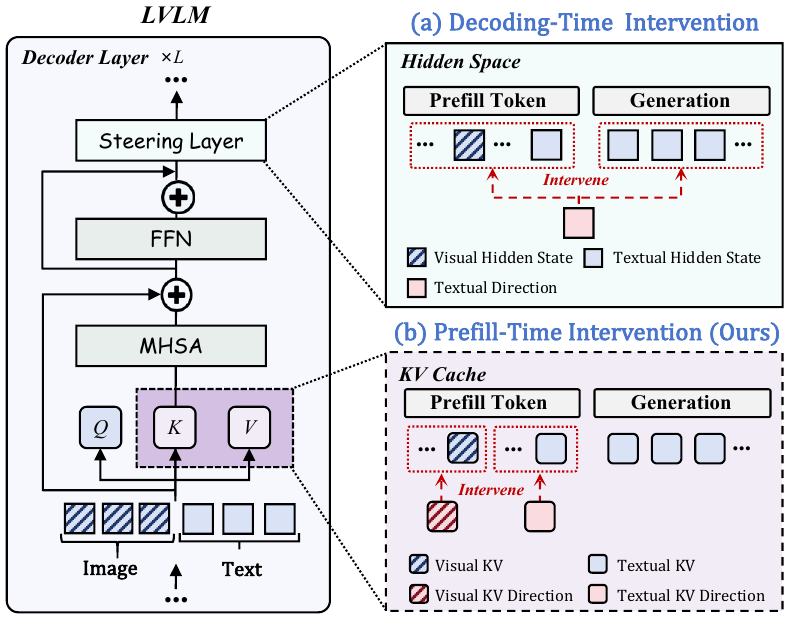} 
    \end{minipage}
   
    \caption{
    Comparative analysis. 
     \textbf{(a):} Decoding-Time Intervention methods continuously intervene in the hidden states of the prefill and generated token. \textbf{(b):} Our method applies modal-specific interventions to the KV cache only once in the prefill phase.
    }
    \label{imgs: intro_2} 
    \vspace{-10pt}
    
\end{figure}

Remarkable advances in large language models (LLMs) \cite{chiang2023vicuna,huang2024good,achiam2023gpt,touvron2023llama2,yang2025qwen3} have driven the emergence of large vision-language models (LVLMs) \cite{liu2023visual,liu2024improved,bai2023qwenvlversatilevisionlanguagemodel,lu2024deepseek,li2025otter,team2025gemma}, which aim to integrate visual perception with linguistic understanding.
Despite their impressive capabilities, LVLMs remain prone to hallucinations \cite{an2025mitigating,suo2025octopus,wu2025antidote,yang2025nullu,yin2025clearsight}, generating factually inconsistent outputs that contradict the visual input. Common manifestations include imaginary entities \cite{rohrbach2018object,li2023evaluating}, incorrect attributes \cite{sun2023aligning,yin2024survey}, and nonexistent relationships \cite{wang2024amberllmfreemultidimensionalbenchmark}.
These failures in factual grounding significantly undermine user trust and impede the safe deployment of LVLMs in real-world interactive applications \cite{chen2024driving,yu2024hallucidoctor,liu2023mitigating}.

\begin{figure}[!th] 
    \centering 

    \begin{minipage}[t]{\linewidth} 
        \centering
        \includegraphics[width=0.95\linewidth]{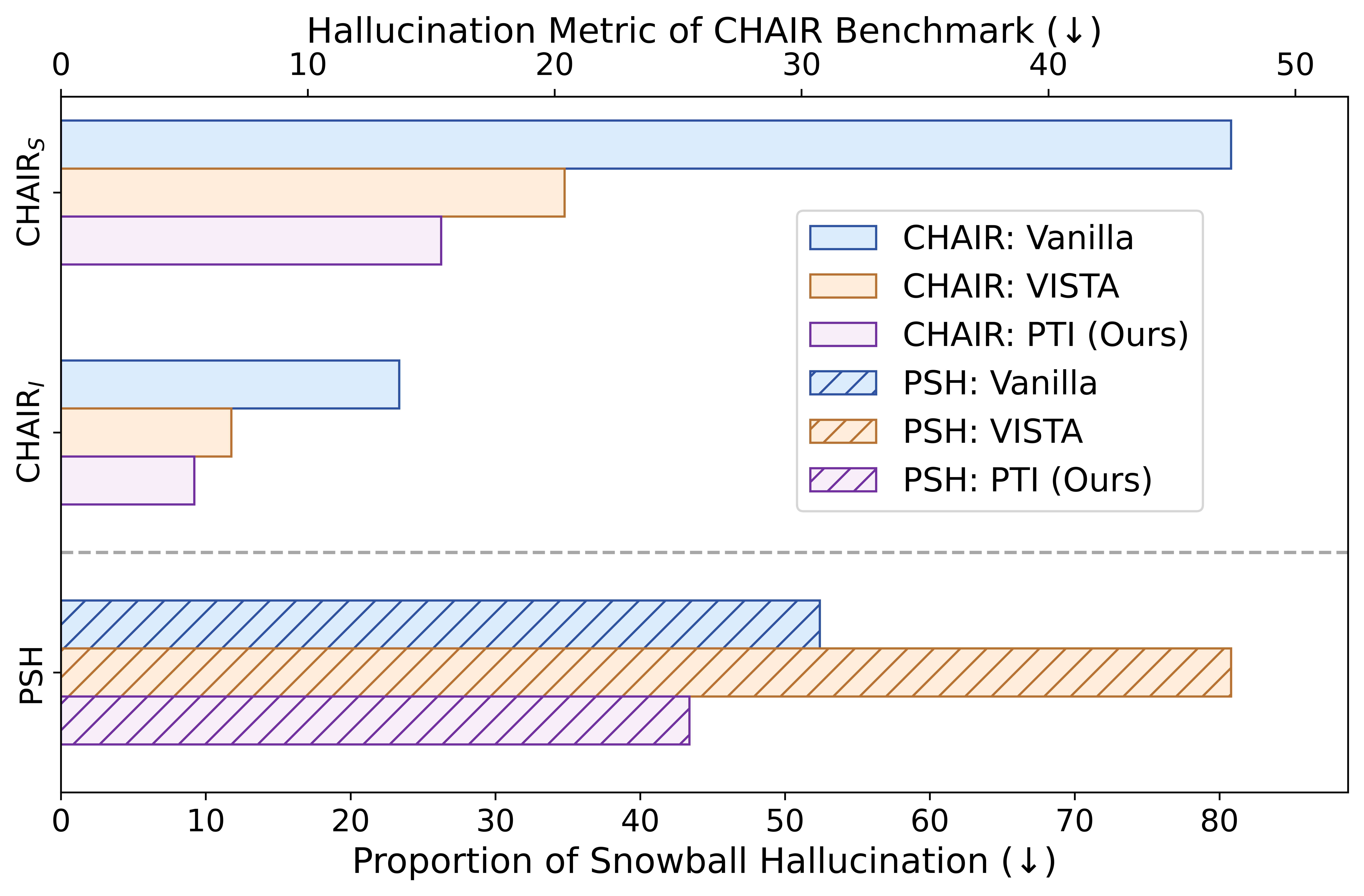} 
    \end{minipage}
    
     \caption{
        Quantitative analysis of LLAVA-1.5 on CHAIR Benchmark \cite{rohrbach2018object}. 
         We report $\text{CHAIR}_S$/$\text{CHAIR}_I$ to measure hallucination frequency at sentence/instance levels, and PSH \cite{tang2025seeing} (the proportion of snowball hallucination, \textit{i.e.}, $\frac{\text{Snowball Hallucinations}}{\text{Overall Hallucinations}}\times100\%$) to quantify the severity of cascading hallucinations.
        }
        \label{imgs: analysis} 
        \vspace{-10pt}
        
\end{figure}

To mitigate these issues, recent approaches adopt the Decoding-Time Intervention (DTI) paradigm \cite{yang2025improving,liu2024reducing,li2025hidden,chen2025ict,yang2025nullu}, demonstrating strong potential for steering LVLM behavior without parameter modification. 
These methods typically apply uniform steering vectors to model representations (\textit{e.g.} hidden states) during decoding (Figure \ref{imgs: intro_2}(a)).
However, a notable challenge remains: although DTI methods such as VISTA \cite{li2025hidden} reduce the frequency of hallucinations, they inadvertently exacerbate the severity of residual hallucinations.
This exacerbation manifests as ``snowball hallucinations'' \cite{tang2025seeing,zhang2024how}: as Figure \ref{imgs: analysis} illustrates, once an initial error is generated, the continuous intervention struggles to halt its propagation, leading to a cascade of progressively degrading inaccuracies. 

We attribute this exacerbation to inherent limitations within the DTI paradigm, rooted in how, what, and when it intervenes.
First, DTI typically employs a uniform steering vector derived solely from the textual state \cite{liu2024reducing,li2025hidden,yang2025nullu}. This modality-agnostic manner ignores the distinct sensitivity of the text decoder to visual representations, potentially worsening the modality misalignment \cite{liu2024reducing} that seeds the initial error.
Furthermore, the intervention targets coarse-grained representation (\textit{e.g.}, high-level hidden states) primarily. 
This lack of precision makes it ill-equipped to correct the fine-grained visual perception errors \cite{chen2025ict,li2025cai}.
Finally, and most critically, the intervention is applied continuously and reactively during the decoding stage. This timing means that it acts after an initial poorly-grounded representation has already been computed, permitting initial errors to accumulate \cite{vogels2025distribution,bi2025llava,li-etal-2025-fairsteer} and snowball \cite{tang2025seeing} autoregressively. 
The solution, therefore, lies not in the decoding stage, but in specific improvements at the source—during the prefill stage when initial representations are formed.

This motivates shifting the focus from continuous decoding-time correction to shaping the model’s initial states, which, in Transformer-based LVLMs, are materialized as the key-value (KV) cache during the prefill stage \cite{tu2024vl,wang-etal-2025-metok}.
Critically, the KV cache is not merely a storage module—it actively shapes every subsequent decoding step by providing contextual information to the attention mechanism \cite{vaswani2017attention}. 
This pivotal role makes it a natural intervention point: recent studies demonstrate that manipulating the KV cache can significantly improve model performance in reasoning \cite{belitsky2025kv,fu2025not,sridhar2025video}, inference acceleration \cite{tu2024vl,wan2025meda,wu2025fast}, and long-context handling \cite{wan2024look,wang2024model}.
Crucially, these gains stem from the pivotal role of the KV cache in shaping the entire subsequent generation process.
We therefore hypothesize that a targeted intervention on the initial KV cache can effectively mitigate hallucinations in LVLMs.

Based on this hypothesis, we propose \textbf{P}refill-\textbf{T}ime \textbf{I}ntervention (\textbf{PTI}), illustrated in Figure \ref{imgs: intro_2} (b). 
Specifically, PTI is structured to directly address the inherent limitations of DTI, adhering to three core principles: \textit{when} to intervene, \textit{how} to intervene, and \textit{what} to target.
First, to rectify DTI's reactive timing, PTI is proactive: it intervenes only once during the prefill stage, before error accumulation can begin. 
Second, to address DTI's modality-agnostic failure, PTI is modality-aware, treating visual and textual inputs distinctly (Figure \ref{imgs:methods}) to precisely counter the modality misalignment that seeds the initial error.
Finally, to solve DTI's lack of precision, PTI targets the fine-grained \textit{KV cache} rather than coarse-grained hidden states. 
This enables direct intervention within the attention mechanism itself \cite{vaswani2017attention}, which naturally acts as a decoupled control mechanism grounded in the distinct roles of keys (where to attend) and values (what to aggregate).
We leverage this property by using object-vs-background contrasts to derive vectors that steer keys toward visually grounded objects and values to filter background noise.
The enhanced cache then serves as a well-grounded initial state for decoding.
 
We experimentally validate PTI on three representative and architecture-distinct LVLMs: LLaVA-1.5 \cite{liu2024improved}, Qwen-VL-Chat \cite{bai2023qwenvlversatilevisionlanguagemodel}, and DeepSeek-VL-Chat \cite{lu2024deepseek}.
PTI achieves superior performance on widely adopted object hallucination and comprehensive benchmarks, outperforming existing decoding-time studies.
In addition, PTI is orthogonal to existing decoding-time methods, enabling seamless plug-and-play integration to further boost their performance.

Our contributions can be summarized as follows.
\begin{itemize}
    \item  We propose \textbf{P}refill-\textbf{T}ime \textbf{I}ntervention (\textbf{PTI}), a novel, plug-and-play, and multi-modal steering paradigm for mitigating hallucinations in LVLMs. 
    \item  Distinct from existing DTI methods, PTI derives distinct directions from the multi-modal cache and decouples them to simultaneously enhance object-centric attention and improve the robustness to background noise. 
    \item Extensive experiments demonstrate that PTI provides a general and robust solution to the existing dilemma of \textit{how}, \textit{what}, and \textit{when} to intervene.
\end{itemize}

\section{Related Work}
\label{sec:formatting}


\subsection{Large Vision-Language Models}

The predominant architecture for modern LVLMs integrates an open-source LLM backbone (\textit{e.g.}, LLAMA \cite{touvron2023llama2}) with a visual encoder using a connector module that projects visual tokens into the LLM's embedding space. 
Most representatively, LLaVA \cite{liu2023visual} utilizes a minimal architecture that connects a standard CLIP \cite{radford2021learning} to its LLM via a simple MLP.
Recent models introduce more sophisticated visual processing.
Qwen-VL-Chat \cite{bai2023qwenvlversatilevisionlanguagemodel} employs a cross-attention resampler to condense variable-length features from its ViT \cite{dosovitskiy2020image} into a fixed-length sequence for the LLM, while DeepSeek-VL-Chat \cite{lu2024deepseek} instead uses a hybrid vision encoder, fusing features from parallel high- and low-resolution encoders \cite{zhai2023sigmoid,kirillov2023segment} to enhance detail-oriented understanding.
Through subsequent modality alignment and instruction tuning, these models exhibit remarkable capabilities on a wide array of visual-language tasks \cite{rohrbach2018object,li2023evaluating,wang2024amberllmfreemultidimensionalbenchmark}.
Despite their remarkable capabilities, the increased complexity and deployment of LVLMs have also exposed them to various security threats and vulnerabilities, motivating the exploration of robust mitigation strategies.

\subsection{Mitigating Hallucination in LVLMs}
Existing studies for mitigating hallucination in LVLMs can be broadly categorized into two types:
1) Visual contrastive decoding \cite{leng2024mitigating,wang2024mitigating,suo2025octopus,tang2025seeing,yin2025clearsight}. These methods contrast model logits derived from original and distorted visual inputs across decoding time steps.
However, the multi-round decoding significantly increases the inference latency. Moreover, recent works \cite{suo2025octopus,tang2025seeing,yin2025clearsight,chen2025ict} have argued that they might discard beneficial language priors and compromise the coherence of generated content. 
2) Decoding-time Intervention (DTI) \cite{li2023inference,liu2024reducing,li2025hidden,chen2025ict}. 
These methods leverage contrastive input samples to extract steering vectors, which are then applied to guide the model's behavior toward specific directions.
A key limitation is that they use a fixed steering vector across all token positions during the overall generation phase, which may exacerbate initial errors and worsen subsequent hallucinatory output.
Although VTI \cite{liu2024reducing} attempts an additional intervention in the visual encoder, its intervention on the linguistically-dominant decoder remains undifferentiated, failing to distinguish between modalities and intervention timing. 
In contrast, our method directly enhances the initial multi-modal KV cache to improve the fidelity of subsequent decoding.

There are two concurrent studies \cite{liu2024deliberation, belitsky2025kv} mostly related to ours.
\cite{liu2024deliberation} trains a differentiable ``coprocessor'' for pre-generation augmentation, while \cite{belitsky2025kv} utilized GPT4o \cite{hurst2024gpt} to transfer controllable reasoning styles to small LLMs.
In contrast, our method is systematically designed around the intervention's modalities and positions to mitigate hallucination in LVLMs. Moreover, our method is more general without re-training or powerful auxiliary models.

\section{Methodology}
\subsection{Preliminaries: Inference with KV Cache}

The standard generative inference process in LVLMs involves two primary stages: 1) prefilling with multi-modal input and 2) decoding with initial KV cache.

\noindent \textbf{Prefilling with Multi-modal Input.} 
During the prefill phase, the model processes the input prompt, which contains sequential visual and textual tokens.
Specifically, this input sequence is represented as a sequence of embeddings $X \in \mathbb{R}^{N_x \times D}$, where $N_x$ is the total sequence length and $D$ is the hidden dimension of the model.

To construct the initial KV cache, this input is processed by each decoder layer $l \in [1, L]$. 
At layer $l$, the input embeddings $X^l \in \mathbb{R}^{N_x \times D}$ are projected via key and value projection matrices $\mathbf{W}^l_K, \mathbf{W}^l_V \in \mathbb{R}^{D \times D}$ as follows:
\begin{equation}
\mathbf{K}^l =X^l \mathbf{W}^l_K, \quad \mathbf{V}^l = X^l \mathbf{W}^l_V.
\end{equation}
The resulting tensors $\mathbf{K}^l, \mathbf{V}^l \in \mathbb{R}^{N_x \times D}$ are then reshaped to $\mathbb{R}^{N_h \times N_x \times d_h}$ to accommodate $N_h$ attention heads, each with dimension $d_h$. 
\textit{For clarity, we refer to these reshaped tensors as $K^l$ and $V^l$ in subsequent steps.}
We denote the structured initial cache as the key set $\mathcal{K}=\{ K^l\}_{l=1}^L$ and value set $\mathcal{V}=\{ V^l\}_{l=1}^L$ across $L$ decoder layers. 
This initial cache stores the layer- and head-level token information, facilitating subsequent token generation.

\noindent\textbf{Decoding with Initial KV Cache.}
\label{Preliminary:decoding}
In the decoding phase, the model generates the output sequence autoregressively, leveraging and updating the KV cache at each step. 

At a given time step $t$ in layer $l$, the new key $\mathbf{k}^{t,l}_{\text{txt}}$ and value $\mathbf{v}^{t,l}_{\text{txt}} $ vectors are computed only for the current input token embeddings $x^{t,l}_{\text{txt}} \in \mathbb{R}^{D}$. 
Simultaneously, the query tensor $\mathbf{q}^{t,l}$ is computed using the query projection matrix $\mathbf{W}^l_Q \in \mathbb{R}^{D \times D}$. 
These new tensors are also reshaped into multi-head format $k^{t,l}_{\text{txt}}, v^{t,l}_{\text{txt}}, q^{t,l}_{\text{txt}} \in \mathbb{R}^{N_h \times 1 \times d_h}$.
The cache at layer $l$ is then updated by concatenating the new key and value vectors. Denoting concatenation by $[\cdot,\cdot]$, the updating process is defined as:
\begin{equation}
    K^l \leftarrow [K^l, k^{t,l}_{\text{txt}}], \quad V^l \leftarrow [V^l, v^{t,l}_{\text{txt}}].
\end{equation}
The model then performs self-attention at layer $l$ using the query and the updated layer cache.
The output token $o^{t,l}_{\text{txt}}  = \text{Softmax}\left(\frac{q^{t,l}_{\text{txt}} (K^l)^T}{\sqrt{d_h}}\right) V^l \in \mathbb{R}^{N_h \times 1 \times d_h}$ is then reshaped back to $\mathbb{R}^D$ and processed by subsequent sub-layers.
Finally, the output of the final decoder layer is then used to predict the next token $x^t_{\text{out}}$.
Our PTI operates on the initial layer-wise caches ($K^l, V^l$) computed during the prefill stage, before the decoding process begins.


\begin{figure*}[!ht] 
    \centering 
    \begin{minipage}[t]{\linewidth} 
        \centering
        \includegraphics[width=\linewidth]{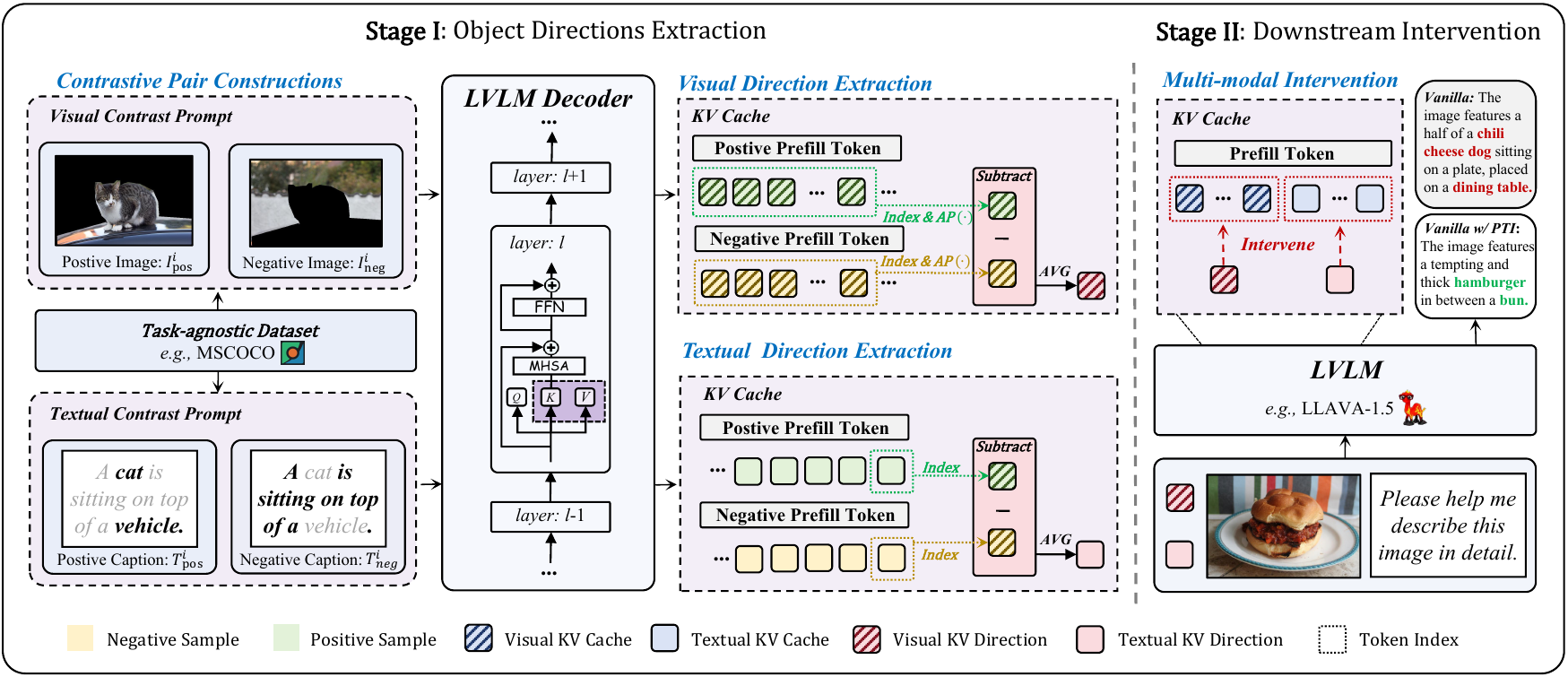} 
    \end{minipage}
   
    \caption{
        Pipeline overview of our PTI. PTI consists of two stages.
        \textbf{Stage I}: 
        we separately extracted the object signals from the visual and textual contrastive KV caches to extract the directions.
        \textbf{Stage II}: 
        these directions are applied as multi-modal interventions to the initial KV cache of downstream input. 
        The modified cache is then passed back to the decoder to generate responses.
    }
    \label{imgs:methods} 
\end{figure*}

\subsection{Object Directions Extraction }

To extract the distinct visual and textual directions, we construct contrastive visual and textual inputs designed to isolate the object signal from the background context.
Since MSCOCO \cite{lin2014microsoft} provides high-quality image-caption pairs and object segments, we utilize this resource to derive task-agnostic directions. 
Notably, the extractions for different modalities in Figure \ref{imgs:methods} (Stage I) operate independently.

\noindent \textbf{Visual Direction Extraction.} 
Given an image $I^i$ and its target object segmentation label $M^i_{\text{obj}}$ from MSCOCO \cite{lin2014microsoft}, we define the object-only signal $I^i_{\text{pos}} = I^i \odot M^i_{\text{obj}}$ as a positive sample. Conversely, the background-only context $I^i_{\text{neg}} = I^i \odot (1 - M^i_{\text{obj}})$ is defined as the negative sample.

Let $\mathcal{I}_{\text{img}}$ denote the indices for all the visual tokens.
Using a fixed textual prompt $T$ to maintain the uniformity of $\mathcal{I}_{\text{img}}$, we perform two separate prefill-time forward passes through the LVLM.
We obtain the positive cache $({K}^{i,l}_{\text{pos}}, {V}^{i,l}_{\text{pos}})$ and the negative cache $({K}^{i,l}_{\text{neg}}, {V}^{i,l}_{\text{neg}})$ at layer $l$ respectively.
Then, we compute the steering vectors from the visual tokens, denoted as:
\begin{equation}
\begin{aligned}
\Delta C^{i,l}_{\text{img}} =  \text{AP}(C^{i,l}_{\text{pos}}- C^{i,l}_{\text{neg}})[\mathcal{I}_{\text{img}}],  \quad  C \in \{ K, V\}
\end{aligned}
\end{equation} 
where $\text{AP}(\cdot)$ denotes the average pooling operation across the visual token dimension \cite{xu2024ote,dosovitskiy2020image}. $S^{i,l}_{\text{k,img}}= \Delta {K}^{i,l}_{\text{img}}$ and $S^{i,l}_{\text{v,img}}=\Delta {V}^{i,l}_{\text{img}}$ represent the visual steering vectors of key and value, respectively. 
Notably, we obtain the final visual directions, ${S}^{l}_{\text{k,img}}$ and ${S}^{l}_{\text{v,img}}$, through averaging over N samples as follows: 
\begin{equation}
\begin{aligned}
{S}^{l}_{\text{k,img}} = \frac{1}{N} \sum_{i=1}^{N} {S}^{i,l}_{\text{k,img}},  \quad   {S}^{l}_{\text{v,img}} = \frac{1}{N} \sum_{i=1}^{N} {S}^{i,l}_{\text{v,img}}.
\end{aligned}
\end{equation}
In addition, we apply PCA \cite{abdi2010principal} via singular value decomposition to remove extra noise following \cite{liu2024reducing,li2025hidden,yang2025nullu}. 
These directions capture the target shift in representation associated with focusing on object signals rather than background context within visual tokens.

\noindent \textbf{Textual Direction Extraction.}  
Inspired by \cite{gan2025textual}, we use a natural language processing tool \cite{vasiliev2020natural} to construct contrastive textual inputs. Specifically, we focus on the visual concept (\textit{e.g.}, ``entity'') associated with target object mentions within the caption.
Consider the corresponding caption $T^i$  from MSCOCO \cite{lin2014microsoft} of the given image $I^i$ (\textit{e.g.}, ``A cat is sitting on top of a vehicle.''), we define $T^i_{\text{pos}}$ as the anchor words that comprise the mentioned objects (\textit{e.g.}, ``cat'' and ``vehicle''). 
Conversely, $T^i_{\text{neg}}$ represents the control set comprising non-anchor words from $T^i$ (\textit{i.e.}, the remaining context after masking the anchor words). 

Similar to the visual extraction process, we perform two separate prefill-time forward passes, while keeping the image constant. Subsequently, we obtain the positive cache $({{\hat K}}^{i,l}_{\text{pos}}, {\hat V}^{i,l}_{\text{pos}})$ and the negative cache $({\hat K}^{i,l}_{\text{neg}}, {\hat V}^{i,l}_{\text{neg}})$ at layer $l$.
Following prior work \cite{liu2024reducing}, we focus on the last textual token in the input sequence, denoting its index as $N_x - 1$. We compute the steering vectors at this token index as,
\begin{equation}
\begin{aligned}
    \Delta \hat C^{i,l}_{\text{txt}} = (\hat C^{i,l}_{\text{pos}}- \hat C^{i,l}_{\text{neg}})[N_x - 1], \quad \hat C \in \{\hat K, \hat V\}
\end{aligned}
\end{equation}
where $ S^{i,l}_{\text{k,txt}}= \Delta {\hat K}^{i,l}_{\text{txt}}$ and $S^{i,l}_{ \text{v,txt}} = \Delta {\hat V}^{i,l}_{\text{txt}}$ represent the textual steering vectors of key and value, respectively. 
We further calculate the averaging textual directions as:
\begin{equation}
\begin{aligned}
    {S}^l_{\text{k,txt}}=\frac{1}{N} \sum_{i=1}^{N} {S}^{i,l}_{\text{k,txt}}, \quad
    {S}^l_{\text{v,txt}}=\frac{1}{N} \sum_{i=1}^{N} {S}^{i,l}_{\text{v,txt}}
\end{aligned}
\end{equation}
and subsequently apply PCA similarly.
These vectors capture the linguistic shift associated with explicitly grounding the text in the object concepts.


\subsection{Downstream Intervention}
During inference on a downstream task, the standard prefill computation yields the initial KV cache for input samples. As shown in Figure \ref{imgs:methods} (Stage II), we inject the extracted visual and textual directions into the corresponding token position of the KV cache of all layers.

\noindent \textbf{Multi-modal Intervention.} 
It is important to clarify that the visual and textual interventions are order-invariant and applied independently. For clarity of presentation, we first detail the intervention on the visual part.
The KV cache (${\tilde{K}^l}, {\tilde{V}^l}$) in layer $l$ at the visual token positions $\mathcal{I}_{\text{img}}$ is shifted as follows:
\begin{equation}
    \begin{aligned}
{\tilde{K}^l}[\mathcal{I}_{\text{img}}] \mathrel{+}= \lambda_{\text{k,img}} {S}^{l}_{\text{k,img}}, \quad
  {\tilde{V}^l}[\mathcal{I}_{\text{img}}] \mathrel{+}=\lambda_{\text{v,img}} {S}^{l}_{\text{v,img}} \\
   \end{aligned}
\end{equation}
where $\lambda_{\text{k,img}}$ and $\lambda_{\text{v,img}}$ denote the scalar coefficients controlling the intervention strength for the keys and values, respectively.
Following \cite{li2025hidden}, we further apply normalization to maintain stability.
This intervention adjusts the initial state of the visual tokens based on the contrast between the object and its background.

Apart from visual cache intervention, we refine the grounding of its linguistic initial state.
Specifically, we apply these textual directions only to the cache entry corresponding to the textual token position $\mathcal{I}_{\text{txt}}$ as follows:
\begin{equation}
\begin{aligned}
{{\tilde{K}}^l}[\mathcal{I}_{\text{txt}}] \mathrel{+}= \lambda_{ \text{k,txt}} {S}^{l}_{\text{k,txt}}, \quad
{{\tilde{V}}^l}[\mathcal{I}_{\text{txt}}] \mathrel{+}= \lambda_{\text{v,txt}} {S}^{l}_{ \text{v,txt}} 
\end{aligned}
\end{equation}
where $\lambda_{\text{k,txt}}$ and $\lambda_{\text{v,txt}}$ control the intervention strength, followed by a similar normalization step.
This adjustment steers the textual representation toward the object-grounded concepts before generation starts.

\noindent \textbf{Model Response Generation.} 
The enhanced initial KV cache serves as the starting state for the autoregressive decoding phase. The model then proceeds with the standard decoding process, as detailed in Section \ref{Preliminary:decoding}. 
Notably, PTI operates only on the initial cache and introduces negligible computational overhead and no modifications during the subsequent decoding steps.

\begin{table*}[!th]
    \vspace{-5pt}
    \centering
    \caption{
    CHAIR hallucination evaluation results across different decoding strategies and LVLMs.
    PTI is compared against SOTA training-free methods that operate during decoding, where ``Vanilla'' stands for the original model. The maximum new token is set to $512$. The best result is highlighted in \textbf{bold}, while the second-best is marked with an \underline{underline}. We provide the latency comparison in Appendix \textcolor{cvprblue}{B}.
    } 
      \vspace{-5pt}
    \label{tab:CHAIR}
    \resizebox{1\linewidth}{!}{
    \begin{tabular}{lccccccc}
        \whline
        \multirow{2}{*}{Decoding}  & \multirow{2}{*}{Method} 
        &\multicolumn{2}{c}{LLAVA-1.5} & \multicolumn{2}{c}{Qwen-VL-Chat} & \multicolumn{2}{c}{DeepSeek-VL-Chat}\\
        \cmidrule(r){3-4} \cmidrule(lr){5-6} \cmidrule(l){7-8}
        
        ~ & & $\text{CHAIR}_S \downarrow$ & $\text{CHAIR}_I \downarrow$ 
          & $\text{CHAIR}_S \downarrow$ & $\text{CHAIR}_I \downarrow$ 
          & $\text{CHAIR}_S \downarrow$ & $\text{CHAIR}_I \downarrow$  \\
        \hline
        \multirow{5}{*}{Greedy} 
 
& \cellcolor{mygray}Vanilla  & \cellcolor{mygray}$47.4$ & \cellcolor{mygray}$13.7$ & \cellcolor{mygray}$39.6$  &  \cellcolor{mygray}$12.0$  &  \cellcolor{mygray}$25.8$  & \cellcolor{mygray}$8.2$  \\
& PAI \cite{liu2024paying} & $22.8$   & $7.0$ & $41.6$  & $12.0$ & $25.6$  & $\bold{6.5}$   \\
& VTI  \cite{liu2024reducing}  & $35.4$ & $11.8$ & $\underline{29.2}$  &  $\underline{8.8}$ &  $24.0$ & $8.4$  \\
& VISTA  \cite{li2025hidden}  & $\underline{20.4} $  & $\underline{6.9}$ & $39.0$  &  $16.8$ & $\underline{22.2}$  &  $7.0$  \\

&  \cellcolor{myblue}\textbf{PTI (Ours)} 
&  \cellcolor{myblue}$\bold{15.4}$ \textcolor{mygreen}{( $\downarrow {32.0}$)}
&  \cellcolor{myblue}$\bold{5.4}$ \textcolor{mygreen}{($\downarrow {8.3}$)}  
&  \cellcolor{myblue}$\bold{20.6}$ \textcolor{mygreen}{($\downarrow {19.0}$)}  
&  \cellcolor{myblue}$\bold{7.0}$ \textcolor{mygreen}{($\downarrow {5.0}$)} 
&  \cellcolor{myblue}$\bold{19.2}$ \textcolor{mygreen}{($\downarrow {6.6}$)} 
&  \cellcolor{myblue}\underline{$6.7$} \textcolor{mygreen}{($\downarrow {1.5}$)}  \\

        \hline
        \multirow{6}{*}{Beam Search}  
 
&  \cellcolor{mygray}Vanilla  & \cellcolor{mygray}$47.8$ & \cellcolor{mygray}$14.2$ & \cellcolor{mygray}$43.6$ &   \cellcolor{mygray}$11.8$  &  \cellcolor{mygray}$27.0$  & \cellcolor{mygray}$7.5$   \\
& PAI   \cite{liu2024paying} & $22.3$ & $6.8$ &  $39.4$ &  $10.6$  & $27.8$  & $6.5$ \\
& OPERA \cite{huang2024opera}  & $45.2$ & $12.4$ & $39.6$  & $11.7$ & $\underline{24.0}$  &  $7.8$  \\
& VTI  \cite{liu2024reducing} & $35.8$ & $11.1$ & $32.2$  & $\underline{8.2}$ & $26.2$  &  $7.2$   \\
& VISTA  \cite{li2025hidden}  & $\underline{17.4}$ & $\underline{6.3}$ & $\underline{30.0}$  &  $12.0$ & $\underline{24.0}$ & $\underline{6.3}$  \\

&  \cellcolor{myblue}\textbf{PTI (Ours)}
&  \cellcolor{myblue}$\bold{13.2}$ \textcolor{mygreen}{($\downarrow {34.6}$)}
&  \cellcolor{myblue}$\bold{5.5}$ \textcolor{mygreen}{($\downarrow  {8.7}$)}  
&  \cellcolor{myblue}$\bold{18.8}$ \textcolor{mygreen}{($\downarrow  {24.8}$)}  
&  \cellcolor{myblue}$\bold{6.3}$ \textcolor{mygreen}{($\downarrow  {5.5}$)} 
&  \cellcolor{myblue}$\bold{15.6}$ \textcolor{mygreen}{($\downarrow  {11.4}$)} 
&  \cellcolor{myblue}$\bold{4.8}$ \textcolor{mygreen}{($\downarrow {2.7}$)}  \\

        \hline
        \multirow{6}{*}{Nucleus Sampling}

&  \cellcolor{mygray}Vanilla  & \cellcolor{mygray}$50.2$ & \cellcolor{mygray}$16.2$ & \cellcolor{mygray}$42.0$  &  \cellcolor{mygray}$14.9$  & \cellcolor{mygray}$34.0$ & \cellcolor{mygray}$11.4$  \\
& PAI  \cite{liu2024paying} &  $43.4$ &  $14.7$ &  $41.6$ &  $12.4$ & $32.2$ & $8.2$  \\
& VCD  \cite{leng2024mitigating} & $51.6$  & $15.3$ & $42.8$ &  $13.3$ & $28.0$  & $8.2$  \\
& VTI  \cite{liu2024reducing}  & $42.6$ & $14.5$ & $\underline{30.4}$  & $\underline{10.0}$ &  $28.6$ & $\underline{10.0}$ \\
& VISTA \cite{li2025hidden}  & $\underline{29.8}$ & $\underline{10.6}$ & $42.8$  &  $17.6$  &  $\underline{22.2}$  & $10.2$  \\

&  \cellcolor{myblue}\textbf{PTI (Ours)} 
&  \cellcolor{myblue}$\bold{25.8}$ \textcolor{mygreen}{($\downarrow  {24.4}$)}
&  \cellcolor{myblue}$\bold{9.2}$ \textcolor{mygreen}{($\downarrow  {7.0}$)}  
&  \cellcolor{myblue}$\bold{18.2}$ \textcolor{mygreen}{($\downarrow  {23.8}$)}  
&  \cellcolor{myblue}$\bold{9.1}$ \textcolor{mygreen}{($\downarrow  { 5.8}$)} 
&  \cellcolor{myblue}$\underline{27.0}$ \textcolor{mygreen}{($\downarrow {7.0}$)} 
&  \cellcolor{myblue}$\bold{8.0}$ \textcolor{mygreen}{($\downarrow  {3.4}$)}  \\

        \whline
        \end{tabular}
    }
\end{table*}

\begin{table*}[!th]
    \vspace{-5pt}
    \centering
    \caption{Experiment results on POPE Benchmark. We report the average accuracy and F1-score computed across three object splits, as well as the specific results on the most challenging adversarial split. We set the maximum new token to 32 and use the nucleus sampling strategy. The complete table can be found in Appendix \textcolor{cvprblue}{B}. 
    }
      \vspace{-5pt}
    \label{tab:POPE}
    \resizebox{1\linewidth}{!}{
    \begin{tabular}{lcccccccccccc}
        \whline
        \multirow{3}{*}{Method } 
        &\multicolumn{4}{c}{LLAVA-1.5} & \multicolumn{4}{c}{Qwen-VL-Chat} & \multicolumn{4}{c}{DeepSeek-VL-Chat}\\
        \cmidrule(r){2-5} \cmidrule(lr){6-9} \cmidrule(l){10-13} 
        & \multicolumn{2}{c}{Adversarial} & \multicolumn{2}{c}{Average} 
        & \multicolumn{2}{c}{Adversarial} & \multicolumn{2}{c}{Average}  
        & \multicolumn{2}{c}{Adversarial} & \multicolumn{2}{c}{Average}  \\
        
        \cmidrule(r){2-3} \cmidrule(lr){4-5} \cmidrule(lr){6-7} \cmidrule(lr){8-9} \cmidrule(lr){10-11} \cmidrule(l){12-13} 

         ~ &   $\text{Acc} \uparrow$  & $\text{F1} \uparrow$  &  $\text{Acc} \uparrow$  & $\text{F1} \uparrow$  &  $\text{Acc} \uparrow$  & $\text{F1} \uparrow$   &  $\text{Acc} \uparrow$  & $\text{F1} \uparrow$ &  $\text{Acc} \uparrow$  & $\text{F1} \uparrow$ &  $\text{Acc} \uparrow$  & $\text{F1} \uparrow$  \\

        \hline

  \cellcolor{mygray}Vanilla  & \cellcolor{mygray}$75.40 $ & \cellcolor{mygray}$ 77.67$ & \cellcolor{mygray}$79.88$ & \cellcolor{mygray}$81.23$ & \cellcolor{mygray}$80.26$ & \cellcolor{mygray}$79.76$ &  \cellcolor{mygray}$83.69$  &  \cellcolor{mygray}$82.92$  & \cellcolor{mygray}$82.73$ & \cellcolor{mygray}$81.97$ & \cellcolor{mygray}$83.89$ & \cellcolor{mygray}$83.06$  \\
 PAI \cite{liu2024paying} & $\underline{76.93}$ & $\bold{79.13}$ &  $\underline{81.73}$ &  $\bold{82.95}$ & $81.63$ & $81.12$ &  $84.91$ &  $84.15$   & $\underline{84.03}$ & $83.23$ & $\underline{85.57}$ & $\underline{84.82}$  \\
VCD  \cite{leng2024mitigating}   & $76.63$ & $\underline{79.01}$ & $81.53$  & $82.80$ & $ \underline{83.10}$ &$ \underline{82.04}$ &  $\underline{85.06}$ &  $\underline{83.72}$ & $83.86$  & $82.89$ &  $85.45$  & $84.47$  \\
 VTI  \cite{liu2024reducing}   & $76.33$ &$ 78.47$ & $80.41$ & $81.64$ & $80.17$  & $79.29$ &  $83.27$  &   $82.19$ &  $83.46$ & $\underline{83.48}$ & $84.81$ & $84.65$ \\
 VISTA  \cite{li2025hidden}  & $75.53$  & $78.61$ & $80.08$ & $81.98$ & $80.80$  & $79.46$ & $83.54$  & $82.12$  & $ \bold{84.80}$ & $83.39$ &  $\bold{85.88}$  & $84.51$  \\


  \cellcolor{myblue}\textbf{PTI (Ours)} 
&  \cellcolor{myblue}$\bold{77.40}$  
&  \cellcolor{myblue}$78.75$  
&  \cellcolor{myblue}$\bold{82.21}$  
&  \cellcolor{myblue}$\underline{82.85}$  
&  \cellcolor{myblue}$\bold{83.37}$  
&  \cellcolor{myblue}$\bold{82.41}$  
&  \cellcolor{myblue}$\bold{85.69}$  
&  \cellcolor{myblue}$\bold{84.62}$ 
&  \cellcolor{myblue}$83.77$  
&  \cellcolor{myblue}$\bold{83.63}$  
&  \cellcolor{myblue}${85.14}$  
&  \cellcolor{myblue}$\bold{85.01}$    \\

        \whline
        \end{tabular}
    }
\end{table*}

\section{Experiments}

\subsection{Experimental Setup}

\noindent\textbf{Model Architectures.} 
We evaluate PTI on three representative LVLMs with distinct architectures: {LLaVA-1.5} \cite{liu2024improved}, {Qwen-VL-Chat} \cite{bai2023qwenvlversatilevisionlanguagemodel}, and {DeepSeek-VL-Chat} \cite{lu2024deepseek}. 

\noindent\textbf{Decoding strategies.} To demonstrate the sustained benefits of PTI, we verify it across three widely used decoding strategies: 
1) {Greedy}, which deterministically selects the most probable token at each step.
2) {Beam Search} with a beam size of $5$. 
3) {Nucleus Sampling} with top-p=$1.0$. 
The temperature is fixed at $1.0$ for all scenarios.

\noindent\textbf{Baselines.} 
We compared PTI with several state-of-the-art (SOTA) training-free hallucination mitigation methods.
For baselines optimizing decoding strategy, we adopt {VCD} \cite{leng2024mitigating} and {OPERA} \cite{huang2024opera}.
For the other decoding-time solutions: we adopt {PAI} \cite{liu2024paying}, {VTI} \cite{liu2024reducing}, and {VISTA} \cite{li2025hidden}.
We reproduce all baseline results using identical evaluation data and settings (\textit{e.g.}, prompt, temperature). 
\textit{To prevent implementation bias, we omitted methods lacking official support for code implementation and specific decoding strategies.}


\subsection{Implementation Details} 
\label{exp:Implementation}
We extract multi-modal directions on a holdout training set containing 100 randomly selected VQA pairs from the MSCOCO \cite{lin2014microsoft}. To assess the effectiveness and robustness of PTI, we evaluated it on Object Hallucination Benchmarks containing CHAIR \cite{rohrbach2018object}, POPE \cite{li2023evaluating}, and AMBER \cite{wang2024amberllmfreemultidimensionalbenchmark}, as well as the comprehensive benchmarks including MMHAL \cite{sun2023aligning} and MME \cite{yin2024survey} with significant distributional differences.   
In the experiment, we set $\lambda_{ \text{k,img}}$ = $\lambda_{ \text{k,txt}}$ and $\lambda_{ \text{v,img}}$ = $\lambda_{ \text{v,txt}}$, and determine the optimal values through a grid search.
Detailed hyperparameter configurations are provided in Appendix \textcolor{cvprblue}{B}.


\subsection{Results on Object Hallucination Benchmarks}
\noindent\textbf{CHAIR Evaluation.}
CHAIR \cite{rohrbach2018object} is a standard metric that quantifies object hallucinations by comparing objects mentioned in generated descriptions against the ground-truth object annotations.
Following \cite{li2025hidden}, We randomly select $500$ images from the MSCOCO validation set with the prompt
``\texttt{Please help me describe this image in detail.}''.
CHAIR comprises two evaluation dimensions:
instance-level 
$\text{CHAIR}_I=\frac{|\{\text{hallucinated object}\}|}{|\{\text{object}\}|}$
and sentence-level
$\text{CHAIR}_S=\frac{|\{\text{caption w/ hallucinated objects}\}|}{|\{\text{caption}\}|}$.
As shown in Table \ref{tab:CHAIR}, distinct from VCD \cite{leng2024mitigating} and OPERA \cite{huang2024opera}, which support specific decoding strategies with minor improvements, PTI demonstrates strong generalization across various decoding strategies. 
Moreover, PTI outperforms both VTI \cite{liu2024reducing} and VISTA \cite{li2025hidden} in most decoding strategies and LVLMs. 
We attribute this robust performance to the modality-specific and one-time intervention design, which brings fine-grained enhancement to the initial representation at the very onset of the decoding phase.
Furthermore, we provide specific case studies in Appendix \textcolor{cvprblue}{C}.

\noindent\textbf{POPE Evaluation.} 
\label{exp:pope}
POPE \cite{li2023evaluating} quantitatively assesses object hallucination by posing targeted, closed-ended questions of the form 
``\texttt{Is there a <object> in the image?}''.
The benchmark comprises three distinct splits of increasing difficulty designed to probe model robustness: \textit{Random}, \textit{Popular}, and \textit{Adversarial}. 
Each split comprises $3000$ VQA pairs drawing from the MSCOCO validation set and is framed as a binary classification task (``\texttt{Yes}''/``\texttt{No}'').
We report the accuracy (ACC) and F1-score (F1) metrics.
The results in Table \ref{tab:POPE} demonstrate PTI’s consistent superiority across LVLMs. Notably, while VTI \cite{liu2024reducing} employs an additional visual intervention to the encoder, its improvements are marginal. For Qwen-VL-Chat, it even led to a decrease from $82.92\%$ to $82.19\%$ ($-0.73\%$) in F1-score.
This degradation is also observed with VISTA ($-0.80\%$), which indicates the necessity of modality-specific differentiation within the dominant decoder.

\begin{table}[ht]
    \vspace{-5pt}
    \centering
    \caption{Experiment results on AMBER Benchmark. 
    The generative task's parameters and metrics follow CHAIR, while the discriminative task's align with POPE.
    }
      \vspace{-5pt}
    \label{tab:Amber_2}
    \resizebox{1\linewidth}{!}{
    \begin{tabular}{lccccc}
        \whline
        \multirow{2}{*}{Model}  & \multirow{2}{*}{Method} 
        &\multicolumn{2}{c}{Greedy} & \multicolumn{2}{c}{Sampling} \\


        \cmidrule(r){3-4} \cmidrule(lr){5-6}
        ~ & & $\text{C}_I \downarrow$ & $\text{Acc} \uparrow $ 
          & $\text{C}_I \downarrow$ & $\text{Acc} \uparrow $ \\
          
        \hline
        \multirow{2}{*}{LLAVA-1.5} 

& \cellcolor{mygray}Vanilla  &  \cellcolor{mygray}$6.1$  & \cellcolor{mygray}$81.5$ &    \cellcolor{mygray}$9.9$ &  \cellcolor{mygray}$73.5$ \\
& \textbf{PTI} & $\bold{3.8}$ & $\bold{82.0}$  &  $\bold{7.3}$ & $\bold{74.6}$ \\

        \hline
        \multirow{2}{*}{Qwen-VL-Chat} 
& \cellcolor{mygray}Vanilla   &   \cellcolor{mygray}$6.5$ &  \cellcolor{mygray}$84.0$ &   \cellcolor{mygray}$8.0$ &  \cellcolor{mygray}$79.9$ \\
& \textbf{PTI} &  $\bold{5.7}$ &  $\bold{84.6}$ &  $\bold{7.4}$ & $\bold{80.9}$ \\

        \hline
        \multirow{2}{*}{DeepSeek-VL-Chat} 

& \cellcolor{mygray}Vanilla  & \cellcolor{mygray}$6.0$ &  \cellcolor{mygray}$85.2$ &   \cellcolor{mygray}$8.1$ &  \cellcolor{mygray}$83.7$ \\

& \textbf{PTI} & $\bold{4.0}$  & $\bold{85.4 }$ & $ \bold{7.0}$ & $\bold{84.1}$ \\

        \whline
        \end{tabular}
    }
      \vspace{-10pt}
\end{table}

\begin{figure*}[!ht]
\centering
\begin{minipage}[t]{0.24\textwidth}
\centering
\vspace{-3pt}
\includegraphics[width=\linewidth]{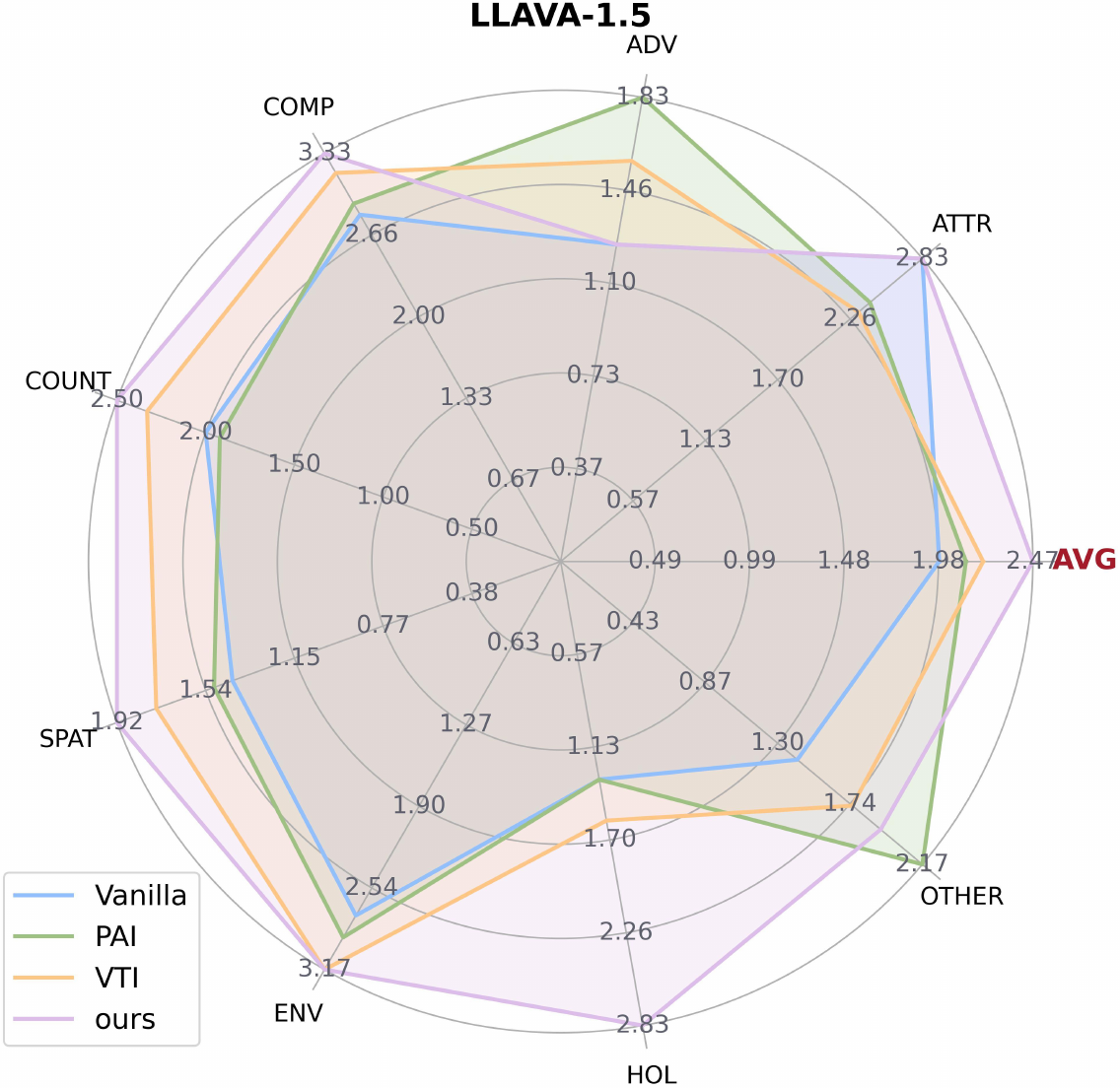}
\vspace{-3pt}
\end{minipage}
\begin{minipage}[t]{0.24\textwidth}
\centering
\vspace{-3pt}
\includegraphics[width=\linewidth]{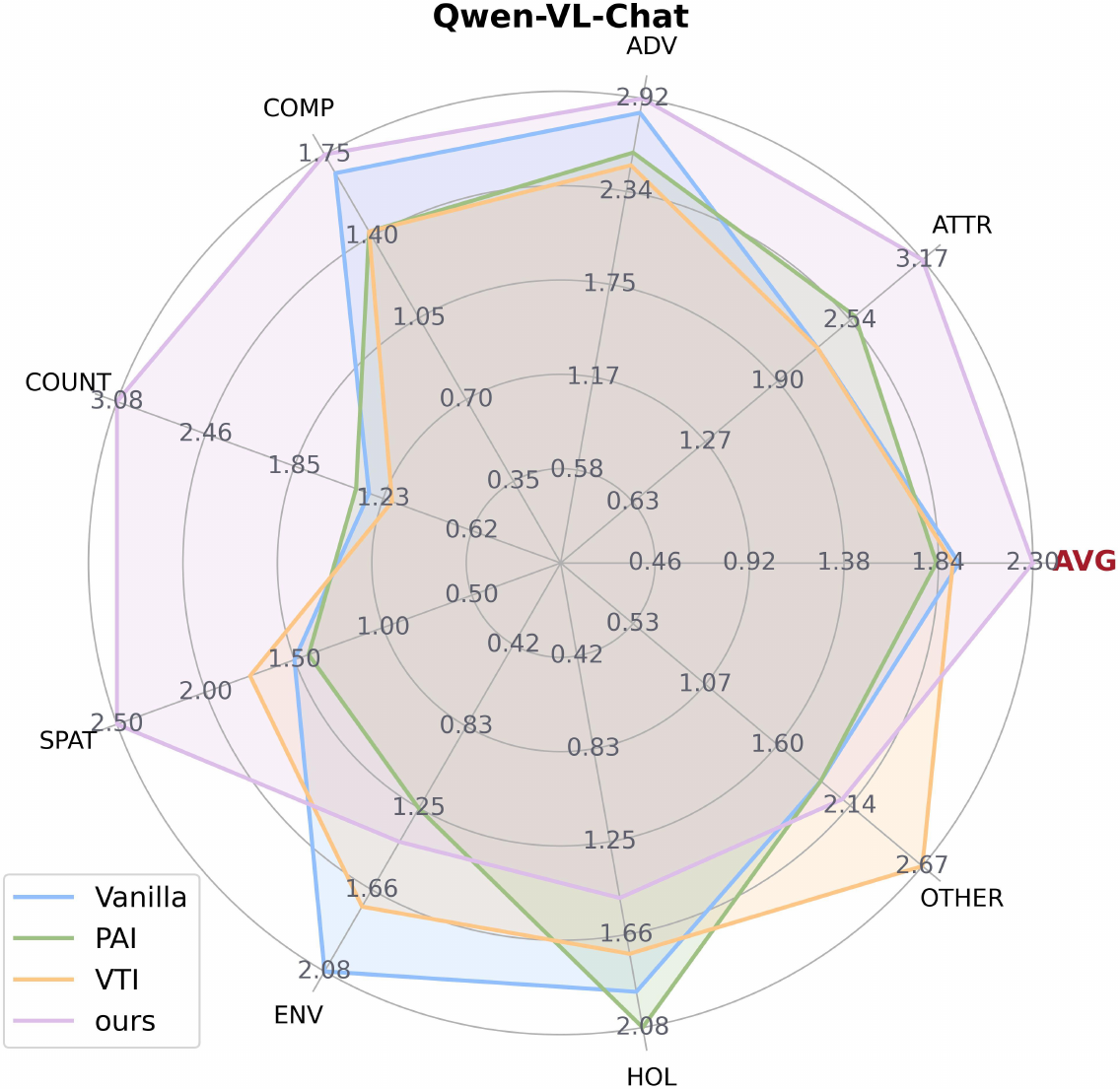}
\vspace{-3pt}
\end{minipage}%
\begin{minipage}[t]{0.24\textwidth}
\centering
\vspace{-3pt}
\includegraphics[width=\linewidth]{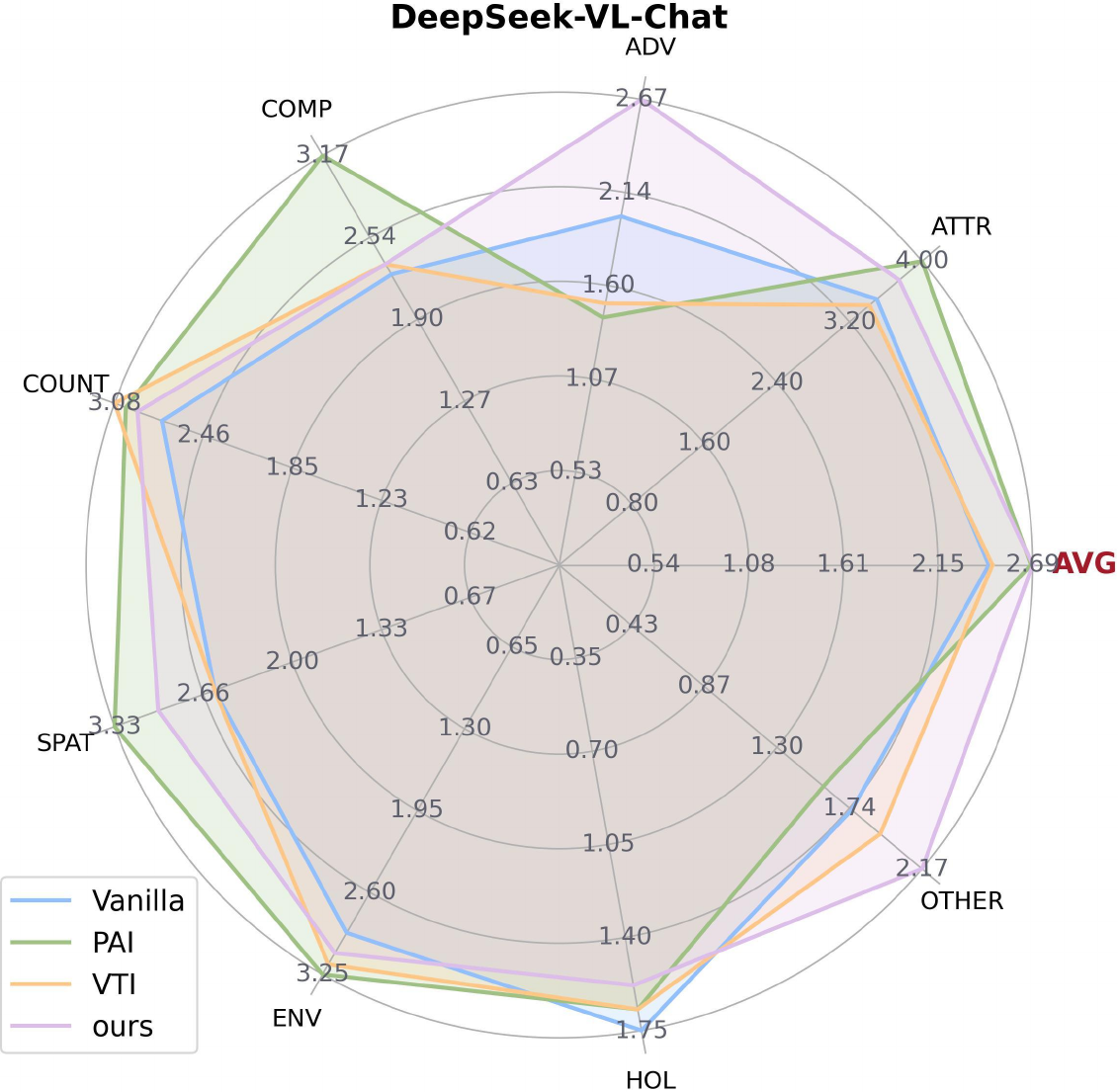}
\vspace{-3pt}
\end{minipage}%
\begin{minipage}[t]{0.24\textwidth}
\centering
\vspace{-3pt}
\includegraphics[width=\linewidth]{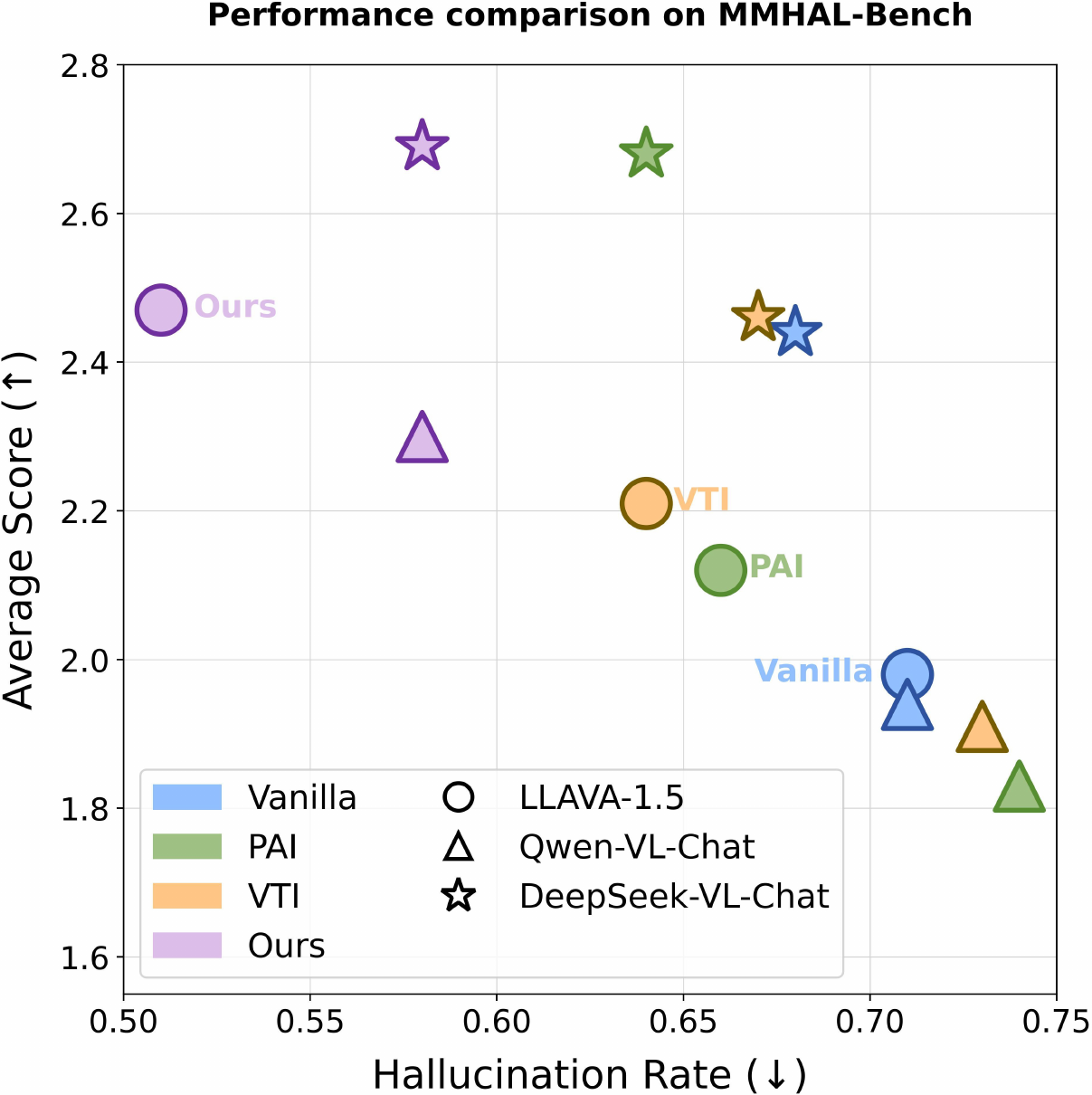}
\vspace{-3pt}
\end{minipage}%
    \vspace{-15pt}
\caption{Performance comparison on MMHal-Bench, with results disaggregated by its eight question categories: attributes (ATTR), adversarial objects (ADV), comparisons (COMP), counting (COUNT), spatial relations (SPAT), environmental inference (ENV), holistic descriptions (HOL), and others (OTHER). All the model responses are evaluated using GPT-5 for alignment with ground-truth answers.}
\label{tab:mmhal}
    \vspace{-5pt}
\end{figure*}

\noindent\textbf{AMBER Evaluation.} 
AMBER \cite{wang2024amberllmfreemultidimensionalbenchmark} is a human-annotated benchmark collected from the MSCOCO test set and ``UnSplash'' website.
The benchmark comprises a generative task with $1,004$ questions and a discriminative task with $14,216$ questions, evaluated by $\text{CHAIR}_I$ ($\text{C}_I$) and accuracy (ACC), respectively. All these questions are designed to assess three categories of hallucination: existence, attributes, and relations.
As shown in Table \ref{tab:Amber_2}, PTI outperforms the vanilla decoding strategies across all LVLMs consistently. 
These results demonstrate the effectiveness of PTI in mitigating object hallucinations and enhancing the model's discriminative understanding ability.


\subsection{Results on Comprehensive Benchmarks}
\noindent\textbf{MMHAL Evaluation.} MMHal-Bench \cite{sun2023aligning} is a specialized benchmark that assesses LVLM hallucinations using $96$ meticulously crafted image-question pairs spanning $8$ distinct categories. It emphasizes logical reasoning and complex visual understanding, and uses GPT-assistance to rate the model's responses. Figure \ref{tab:mmhal} presents the experimental results comparing PTI with PAI \cite{liu2024paying} and VTI \cite{liu2024reducing} based on greedy decoding.
Evidently, PTI delivers significant gains in average scores while concurrently achieving substantial reductions in hallucination rates. 
These improvements are particularly pronounced for the LLaVA-1.5 and Qwen-VL-Chat. 
Furthermore, we observe that PTI exceptionally excels in challenging tasks, such as counting (COUNT), spatial relations (SPAT), and attributes (ATTR). 
This is attributed to the fact that these categories fundamentally require a fine-grained capture of object details, which is essential for accurate responses.
In summary, these results demonstrate the effectiveness of PTI in mitigating a broader range of hallucination challenges.


\begin{table}[ht]
    \vspace{-5pt}
    \centering
    \caption{Evaluation of MME Benchmark. 
    ``LLAVA.'': LLAVA-1.5, ``Qwen.'': Qwen-VL-Chat, ``DeepSeek.'': DeepSeek-VL-Chat.
    }
      \vspace{-5pt}
    \label{tab:mme}
    \resizebox{1\linewidth}{!}{
    \begin{tabular}{lccccccc}
        \whline
         \multirow{2}{*}{Method} 
        & \multicolumn{2}{c}{LLAVA.} &\multicolumn{2}{c}{Qwen.} &\multicolumn{2}{c}{DeepSeek.}\\

        \cmidrule(r){2-3} \cmidrule(r){4-5}  \cmidrule(r){6-7}  
        ~  &  $\text{Acc} \uparrow$ & $ \Delta $  &  $\text{Acc} \uparrow$ & $ \Delta $  &  $\text{Acc} \uparrow$ & $ \Delta $  \\
         \hline




\rowcolor{mygray}
Vanilla &    
\tabincell{c}{$611.6$} &  $-$   &   
\tabincell{c}{$598.3$} &  $-$   & 
\tabincell{c}{$651.6$} &   $-$  \\

PAI \cite{liu2024paying} & 
\tabincell{c}{$625.0$} & \textcolor{mydarkgray}{$\uparrow 13.4$}   & 
\tabincell{c}{$605.0$} & \textcolor{mydarkgray}{$\uparrow 6.7$}   & 
\tabincell{c}{$656.6$} & \textcolor{mydarkgray}{$\uparrow 5.0$}  
\\

VTI \cite{liu2024reducing}  &
\tabincell{c}{$\underline{633.3}$} & \textcolor{mydarkgray}{$\uparrow 21.7$} &
\tabincell{c}{$\underline{626.6}$} & \textcolor{mydarkgray}{$\uparrow 28.3$} & 
\tabincell{c}{$\underline{661.6}$} & \textcolor{mydarkgray}{$\uparrow 10.0$}
\\

VISTA  \cite{li2025hidden}  & 
\tabincell{c}{$615.0$} &  \textcolor{mydarkgray}{$\uparrow 3.4$}  & 
\tabincell{c}{$611.6$} &  \textcolor{mydarkgray}{$\uparrow 13.3$}   & 
\tabincell{c}{$646.6$} &  \textcolor{mydarkgray}{$\downarrow 5.0$} 

\\

\rowcolor{myblue}
\textbf{PTI (ours)} &   
\tabincell{c}{\textbf{$\bold{651.6}$}}  & \textcolor{mygreen}{$\uparrow 40.0$}   & 
\tabincell{c}{\textbf{$\bold{638.3}$}}  & \textcolor{mygreen}{$\uparrow 40.0$}  & 
\tabincell{c}{\textbf{$\bold{671.6}$}} &  \textcolor{mygreen}{$\uparrow 20.0$}  \\

        \whline
        \end{tabular}
    }
      \vspace{-5pt}
\end{table}

\noindent\textbf{MME Evaluation.} 
The MME benchmark \cite{yin2024survey} is designed to assess LVLM performance across multiple dimensions comprehensively. 
We mainly focus on its cognition-related subsets, which specifically address two types of hallucinations: object-level (existence and count) and attribute-level (position and color).
As shown in Table \ref{tab:mme}, we report the performance across all tasks is measured using accuracy, based on greedy decoding. 
Obviously, PTI achieved the best total performance, while VTI \cite{liu2024reducing} also demonstrated competitive results. This significant performance gap relative to other methods underscores the critical role of enhancing visual signals in boosting the model's perceptual capabilities. 
The complete table is detailed in Appendix \textcolor{cvprblue}{B}.

\begin{figure*}[!ht]
\centering

\begin{minipage}[t]{0.34 \textwidth}
\centering
\includegraphics[width=\linewidth]{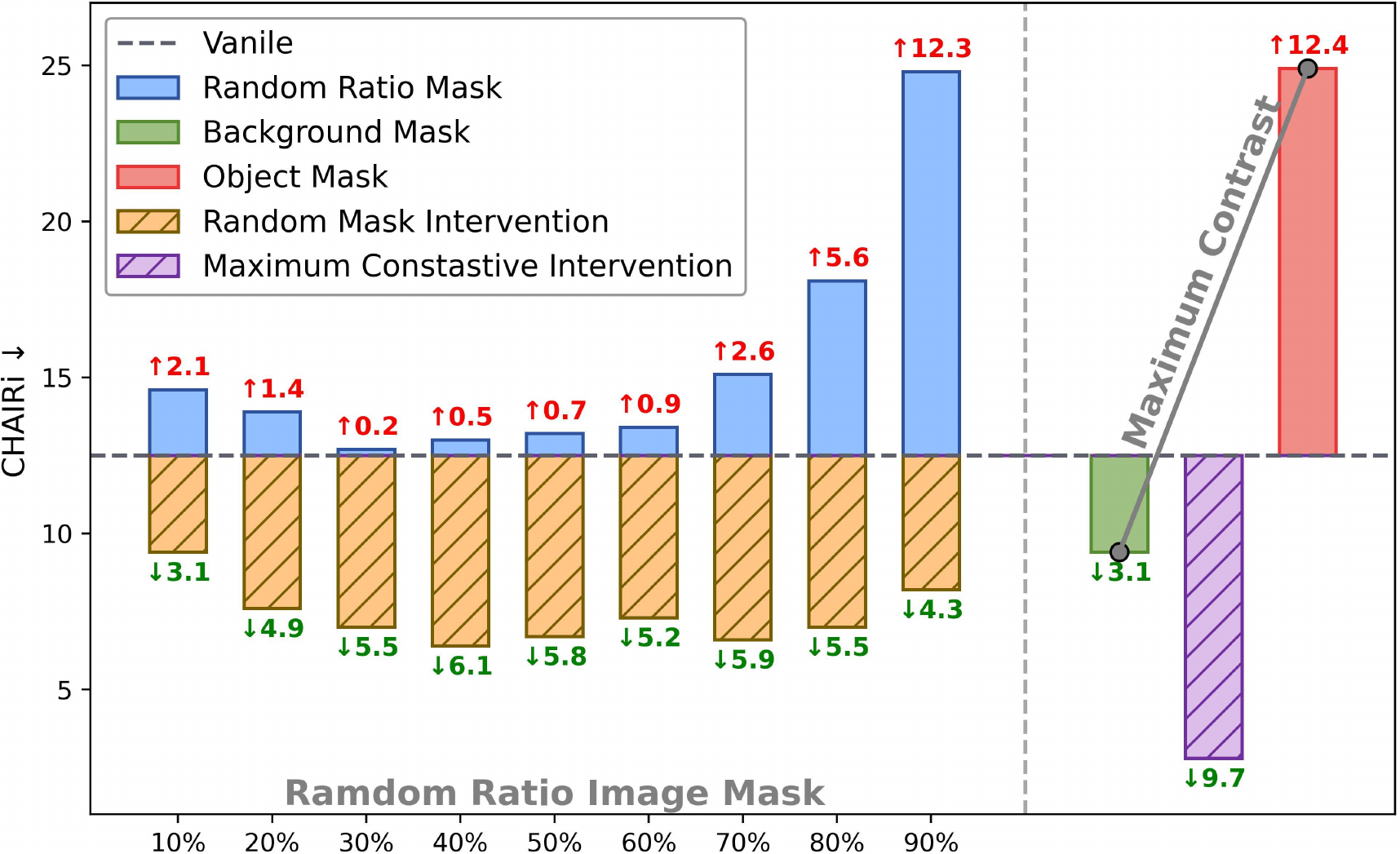}
\subcaption{Visual Value Cache Intervention} 
\end{minipage}%
\hspace{0.02\textwidth} 
\begin{minipage}[t]{0.63 \textwidth} 
    \begin{minipage}[t]{0.56\textwidth}
    \centering
    \includegraphics[width=\linewidth]{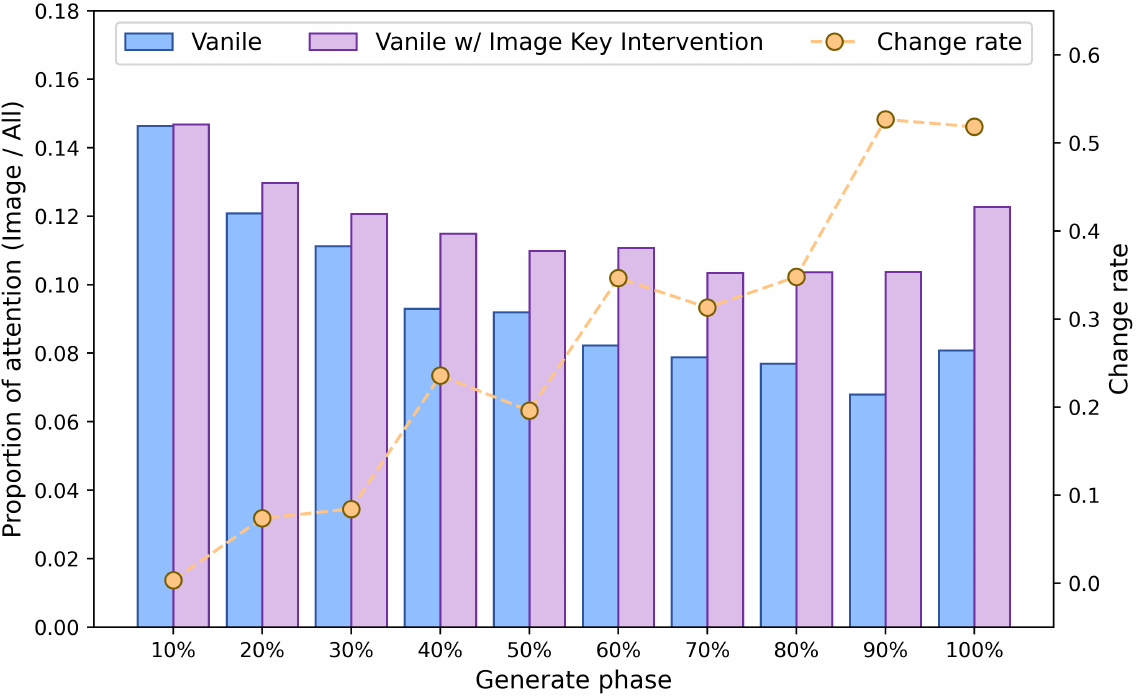}
    \end{minipage}%
    \hspace{0.01\textwidth} 
    \begin{minipage}[t]{0.42 \textwidth}
    \centering
    \includegraphics[width=\linewidth]{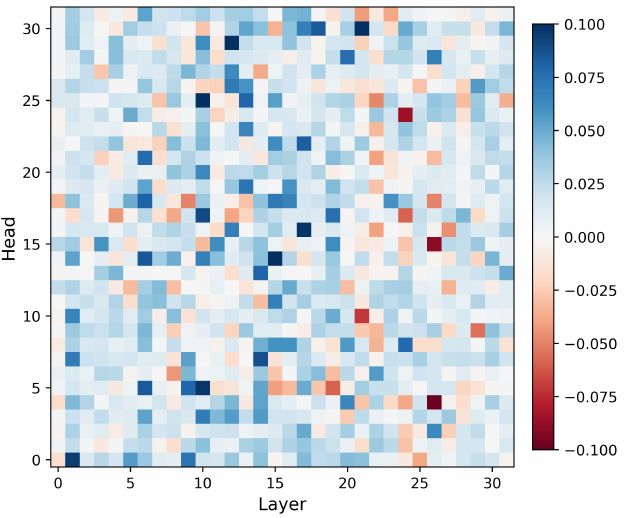}
    \end{minipage}%
\subcaption{Visual Key Cache Intervention} %
\end{minipage}%

    \vspace{-5pt}
\caption{
    Internal interpretability analysis of visual cache intervention on LLAVA-1.5 across 300 randomly selected images from MSCOCO.
    \textbf{(a):} Ablation study on value intervention strategies, validating that the maximal contrast (object vs. background) yields the largest hallucination reduction.
    \textbf{(b):} Analysis of the key cache intervention, demonstrating its dual effect: mitigating global visual attention decay during generation (left) and enhancing local, object-centric attention (right).
    The change rate formula is detailed in Appendix \textcolor{cvprblue}{A}. 
    }
\label{tab:analysis_Interp}
    \vspace{-5pt}
\end{figure*}


\section{Analysis and Discussions}
\label{analysis:total}

\subsection{Ablation Study and Error Analysis}
\label{analysis:Ablation}

To validate our modality-specific intervention design, we conduct ablation studies both on the intervention modalities and positions. 
The results in Table \ref{tab:analysis_ablation} reveal a clear, modality-dependent dichotomy:
Textual intervention is more effective when precise, steering the final generative state, while visual intervention yields optimal results when applied globally to enhance the entire visual initial state.
Specifically, the Visual intervention is the most crucial component for hallucination reduction, driving significant performance improvements. However, it also induced a drop in the F1 score. 
This suggests that an overemphasis on the visual modality may bias the model towards fine-grained details, 
potentially at the expense of degraded generation quality.
In summary, our final PTI method pursues this trade-off. 
The additional precise textual intervention not only complements the visual concepts but also recovers the F1 score.
This confirms that PTI successfully combines these two strategies, achieving the best overall performance.

\begin{table}[ht]
    \vspace{-5pt}
    \centering
    \caption{Ablation study for intervention modality and position on LLAVA-1.5. ``$\text{C}_S$'': $\text{CHAIR}_S$, ``$\text{C}_I$'': $\text{CHAIR}_I$, ``F1'': F1-score. 
    }
      \vspace{-5pt}
    \label{tab:analysis_ablation}
    \resizebox{1\linewidth}{!}{
    \begin{tabular}{ccccccc}
        \whline
        
        \multicolumn{2}{c}{\makebox[0.18\textwidth] {Modality}}   
        &   \multicolumn{2}{c}{\makebox[0.18\textwidth]{Position}}
        & \multirow{2}{*}{$\text{C}_S \downarrow$} 
        & \multirow{2}{*}{$\text{C}_I \downarrow$} 
        & \multirow{2}{*}{$\text{F1} \uparrow$} \\
        
        \cmidrule(r){1-2} \cmidrule(r){3-4}   
        \text{Textual}  &  \text{Visual} &  \text{Last token}  &  \text{All tokens}   \\

         \hline
         
\rowcolor{mygray}
$\times$ & $\times$ & $\times$ & $\times$ & $47.4$ &  $13.7$  & $75.3$    \\

$ \checkmark $ & $\times$ & $\checkmark$ & $\times$
& $40.8$ &  $12.0$  & $76.5$   \\

$ \checkmark $ & $\times$ & $\times$ & $\checkmark$
& $45.2$ &  $ 14.3$  & $75.6$ \\

$ \times $ & $\checkmark$ & $\checkmark$ & $\times$
& $41.2$ &  $12.4$  & $76.4$  \\

$ \times $ & $\checkmark$ & $\times$ & $\checkmark$
& $16.8$ &  $6.2$  & $70.3$  \\

\hline

\rowcolor{myblue}
 \multicolumn{4}{c}{PTI: Last textual token \& All visual tokens}  & $15.4$ & $5.4$ & $72.7$   \\

        \whline
        \end{tabular}
    }
      \vspace{-5pt}
\end{table}

\subsection{Internal Interpretability Analysis}
\label{analysis:Internal}
Given the critical role of visual intervention shown in Section \ref{analysis:Ablation}, we further conduct an internal interpretability analysis.
Specifically, we decouple the interventions on the key and value caches to analyze their respective contributions.

\noindent\textbf{Value Cache Intervention.} 
As shown in the left panel of Figure \ref{tab:analysis_Interp} (a), directly perturbing the original images via random masking leads to a severe exacerbation of hallucinations.
We attribute these results to the fact that random masking increases the risk of losing object signals.
Therefore, a straightforward strategy is to leverage the contrastive directions derived from the original and perturbed images to enhance model robustness. 
However, the efficacy of this strategy appears saturated with no clear differences between masking ratios.
These results motivated us to search for a more principled and potent contrast.
Our intuition is that the contrast between the isolated object and its background yield the maximal object-centric signal. 
The right panel of the Figure \ref{tab:analysis_Interp} (a) validates this point of view, and achieves a maximal hallucination reduction ($-9.7\%$ of $\text{CHAIR}_I$), significantly outperforming all random-mask-based contrast and confirming our methodological choice.

\noindent\textbf{Key Cache Intervention.} 
We further analyzed the impact of our visual key cache intervention in Figure \ref{analysis:total} (b). 
The left part shows that LVLMs' attention proportion allocated to the overall visual tokens progressively decays during generation. 
In contrast, our intervention consistently mitigates this decay.
Critically, the relative change rate demonstrates that the positive impact of our intervention becomes increasingly pronounced as the inference phase progresses.
Beyond this global improvement, we verified that this enhanced visual attention is meaningfully focused. 
The right part visualizes a targeted shift in attention toward object-level details across layers and heads (attention maps are visualized in Appendix \textcolor{cvprblue}{C}).
Overall, these results validate that our intervention not only preserves global visual grounding but also effectively guides attention toward local details.

\begin{table}[!ht]
    \vspace{-5pt}
    \centering
    \caption{Cross-models and Comb-methods generalization study. Performance on POPE \textit{Adversarial} subset. ``$\mathcal{L} \Longleftrightarrow \mathcal{Q}$'': We restrict our cross-model generalization analysis to LLaVA-1.5 and Qwen-VL-Chat, as they share identical KV cache dimensions.}
      \vspace{-5pt}
    \label{tab:analysis_combine}
    \resizebox{1\linewidth}{!}{
    \begin{tabular}{lccccccc}
        \whline
        \multirow{2}{*}{Setting}   & \multirow{2}{*}{Method} 
        &\multicolumn{2}{c}{LLAVA.} &\multicolumn{2}{c}{Qwen.} &\multicolumn{2}{c}{DeepSeek.}\\

        \cmidrule(r){3-4} \cmidrule(r){5-6}  \cmidrule(r){7-8}  
        ~ & &  $\text{Acc.} \uparrow$ & $ \Delta $  &  $\text{Acc.} \uparrow$ & $ \Delta $  &  $\text{Acc.} \uparrow$ & $ \Delta $  \\
         \hline
        \multirow{2}{*}{\makecell{Cross\\LVLMs}}

& \cellcolor{mygray}Vanilla  &  \cellcolor{mygray}$75.40$  & \cellcolor{mygray}$-$   &  \cellcolor{mygray}$80.26$  & \cellcolor{mygray}$-$  &  \cellcolor{mygray}$82.73$  & \cellcolor{mygray}$-$  \\

&  PTI ($\mathcal{L} \Longleftrightarrow \mathcal{Q}$) 
& $75.53$ &  \textcolor{mygreen}{$\uparrow 0.13$}  
& $81.47$ & \textcolor{mygreen}{$\uparrow 1.21$}  
& $-$ & $-$  \\

        \hline
        
        \multirow{4}{*}{\makecell{Integrated\\Methods}} 

& \cellcolor{mygray}PAI  \cite{liu2024paying}
& \cellcolor{mygray}$76.93$ & \cellcolor{mygray}$-$
& \cellcolor{mygray}$81.63$ & \cellcolor{mygray}$-$
& \cellcolor{mygray}$84.03$ & \cellcolor{mygray}$-$ \\

& {PAI \textit{w/} PTI} 
& $78.76$ & \textcolor{mygreen}{$\uparrow 1.83$}  
& {${82.30}$} & \textcolor{mygreen}{$\uparrow 0.67$}  
& {${84.60}$} & \textcolor{mygreen}{$\uparrow 0.57$} \\

& \cellcolor{mygray}VISTA  \cite{li2025hidden}
& \cellcolor{mygray}$75.53$ & \cellcolor{mygray}{$-$} 
& \cellcolor{mygray}$80.80$ & \cellcolor{mygray}{$-$}
& \cellcolor{mygray}$84.80$ & \cellcolor{mygray}{$-$}  \\

& {VISTA \textit{w/} PTI} 
& $76.33$ & \textcolor{mygreen}{$\uparrow 0.80$}  
& $80.96$ & \textcolor{mygreen}{$\uparrow 0.16$} 
& {${85.56}$} & \textcolor{mygreen}{$\uparrow 0.76$}    \\

        \whline
        \end{tabular}
    }
      \vspace{-10pt}
\end{table}

\subsection{Generalizability Analysis}
\noindent\textbf{Generalizability between Different LVLMs.} 
We first evaluate the cross-model generalizability of our PTI to determine if the extracted steering vectors capture model-agnostic mechanisms for mitigating hallucinations. 
As shown in Table \ref{tab:analysis_combine}, applying PTI derived from one to another yields accuracy improvements compared to the vanilla.
This positive transfer suggests that PTI captures representational properties related to objects that possess a degree of universality. 
This finding underscores the potential applicability of PTI beyond the architecture of the source model.

\noindent\textbf{Generalizability integrated with other methods.} 
Furthermore, we investigate the complementarity of PTI by integrating it with other methods operating at different stages. 
Table \ref{tab:analysis_combine} shows consistent and additive performance gains across all three diverse LVLMs.
These synergistic improvements indicate that PTI operates through an orthogonal mechanism distinct from existing methods, highlighting PTI's value not only as a standalone technique but also as a computationally efficient and effective module.

\section{Conclusion}

In this paper, we propose a novel multi-modal steering paradigm, PTI, to mitigate hallucination in LVLMs.
Unlike existing decoding-stage methods, PTI shifts the intervention to the prefill stage to overcome potential progressive error cascades. 
PTI precisely targets the initial KV cache with modality-aware and position-sensitive corrections, enabling decoupled Key and Value interventions to simultaneously enhance object-centric attention and robustness to background noise.
Extensive experiments demonstrate that PTI not only achieves state-of-the-art performance across diverse benchmarks and LVLMs, but also is orthogonal to existing methods.

\section*{Acknowledgements}
This work was supported by the Natural Science Foundation of China under Grant 62571507.

{
    \small
    \bibliographystyle{ieeenat_fullname}
    \bibliography{main}

@String(AAAI = {AAAI})

@inproceedings{liu2024improved,
  title={Improved baselines with visual instruction tuning},
  author={Liu, Haotian and Li, Chunyuan and Li, Yuheng and Lee, Yong Jae},
  booktitle={Proceedings of the IEEE/CVF conference on computer vision and pattern recognition},
  pages={26296--26306},
  year={2024}
}

@misc{bai2023qwenvlversatilevisionlanguagemodel,
      title={Qwen-VL: A Versatile Vision-Language Model for Understanding, Localization, Text Reading, and Beyond}, 
      author={Jinze Bai and Shuai Bai and Shusheng Yang and Shijie Wang and Sinan Tan and Peng Wang and Junyang Lin and Chang Zhou and Jingren Zhou},
      year={2023},
      eprint={2308.12966},
      archivePrefix={arXiv},
      primaryClass={cs.CV},
      url={https://arxiv.org/abs/2308.12966}, 
}

@inproceedings{leng2024mitigating,
  title={Mitigating object hallucinations in large vision-language models through visual contrastive decoding},
  author={Leng, Sicong and Zhang, Hang and Chen, Guanzheng and Li, Xin and Lu, Shijian and Miao, Chunyan and Bing, Lidong},
  booktitle={Proceedings of the IEEE/CVF Conference on Computer Vision and Pattern Recognition},
  pages={13872--13882},
  year={2024}
}

@article{lu2024deepseek,
  title={Deepseek-vl: towards real-world vision-language understanding},
  author={Lu, Haoyu and Liu, Wen and Zhang, Bo and Wang, Bingxuan and Dong, Kai and Liu, Bo and Sun, Jingxiang and Ren, Tongzheng and Li, Zhuoshu and Yang, Hao and others},
  journal={arXiv preprint arXiv:2403.05525},
  year={2024}
}

@inproceedings{radford2021learning,
  title={Learning transferable visual models from natural language supervision},
  author={Radford, Alec and Kim, Jong Wook and Hallacy, Chris and Ramesh, Aditya and Goh, Gabriel and Agarwal, Sandhini and Sastry, Girish and Askell, Amanda and Mishkin, Pamela and Clark, Jack and others},
  booktitle={International conference on machine learning},
  pages={8748--8763},
  year={2021},
  organization={PmLR}
}

@article{dosovitskiy2020image,
  title={An image is worth 16x16 words: Transformers for image recognition at scale},
  author={Dosovitskiy, Alexey},
  journal={arXiv preprint arXiv:2010.11929},
  year={2020}
}

@inproceedings{zhai2023sigmoid,
  title={Sigmoid loss for language image pre-training},
  author={Zhai, Xiaohua and Mustafa, Basil and Kolesnikov, Alexander and Beyer, Lucas},
  booktitle={Proceedings of the IEEE/CVF international conference on computer vision},
  pages={11975--11986},
  year={2023}
}

@inproceedings{kirillov2023segment,
  title={Segment anything},
  author={Kirillov, Alexander and Mintun, Eric and Ravi, Nikhila and Mao, Hanzi and Rolland, Chloe and Gustafson, Laura and Xiao, Tete and Whitehead, Spencer and Berg, Alexander C and Lo, Wan-Yen and others},
  booktitle={Proceedings of the IEEE/CVF international conference on computer vision},
  pages={4015--4026},
  year={2023}
}

@inproceedings{liu2024paying,
  title={Paying more attention to image: A training-free method for alleviating hallucination in lvlms},
  author={Liu, Shi and Zheng, Kecheng and Chen, Wei},
  booktitle={European Conference on Computer Vision},
  pages={125--140},
  year={2024},
  organization={Springer}
}

@article{liu2024reducing,
  title={Reducing hallucinations in vision-language models via latent space steering},
  author={Liu, Sheng and Ye, Haotian and Xing, Lei and Zou, James},
  journal={arXiv preprint arXiv:2410.15778},
  year={2024}
}

@article{li2025hidden,
  title={The hidden life of tokens: Reducing hallucination of large vision-language models via visual information steering},
  author={Li, Zhuowei and Shi, Haizhou and Gao, Yunhe and Liu, Di and Wang, Zhenting and Chen, Yuxiao and Liu, Ting and Zhao, Long and Wang, Hao and Metaxas, Dimitris N},
  journal={arXiv preprint arXiv:2502.03628},
  year={2025}
}

@inproceedings{huang2024opera,
  title={Opera: Alleviating hallucination in multi-modal large language models via over-trust penalty and retrospection-allocation},
  author={Huang, Qidong and Dong, Xiaoyi and Zhang, Pan and Wang, Bin and He, Conghui and Wang, Jiaqi and Lin, Dahua and Zhang, Weiming and Yu, Nenghai},
  booktitle={Proceedings of the IEEE/CVF Conference on Computer Vision and Pattern Recognition},
  pages={13418--13427},
  year={2024}
}

@article{li2023evaluating,
  title={Evaluating object hallucination in large vision-language models},
  author={Li, Yifan and Du, Yifan and Zhou, Kun and Wang, Jinpeng and Zhao, Wayne Xin and Wen, Ji-Rong},
  journal={arXiv preprint arXiv:2305.10355},
  year={2023}
}

@article{rohrbach2018object,
  title={Object hallucination in image captioning},
  author={Rohrbach, Anna and Hendricks, Lisa Anne and Burns, Kaylee and Darrell, Trevor and Saenko, Kate},
  journal={arXiv preprint arXiv:1809.02156},
  year={2018}
}

@misc{wang2024amberllmfreemultidimensionalbenchmark,
      title={AMBER: An LLM-free Multi-dimensional Benchmark for MLLMs Hallucination Evaluation}, 
      author={Junyang Wang and Yuhang Wang and Guohai Xu and Jing Zhang and Yukai Gu and Haitao Jia and Jiaqi Wang and Haiyang Xu and Ming Yan and Ji Zhang and Jitao Sang},
      year={2024},
      eprint={2311.07397},
      archivePrefix={arXiv},
      primaryClass={cs.CL},
      url={https://arxiv.org/abs/2311.07397}, 
}

@inproceedings{lin2014microsoft,
  title={Microsoft coco: Common objects in context},
  author={Lin, Tsung-Yi and Maire, Michael and Belongie, Serge and Hays, James and Perona, Pietro and Ramanan, Deva and Doll{\'a}r, Piotr and Zitnick, C Lawrence},
  booktitle={European conference on computer vision},
  pages={740--755},
  year={2014},
  organization={Springer}
}

@article{sun2023aligning,
  title={Aligning large multimodal models with factually augmented rlhf},
  author={Sun, Zhiqing and Shen, Sheng and Cao, Shengcao and Liu, Haotian and Li, Chunyuan and Shen, Yikang and Gan, Chuang and Gui, Liang-Yan and Wang, Yu-Xiong and Yang, Yiming and others},
  journal={arXiv preprint arXiv:2309.14525},
  year={2023}
}

@article{yin2024survey,
  title={A survey on multimodal large language models},
  author={Yin, Shukang and Fu, Chaoyou and Zhao, Sirui and Li, Ke and Sun, Xing and Xu, Tong and Chen, Enhong},
  journal={National Science Review},
  volume={11},
  number={12},
  pages={nwae403},
  year={2024},
  publisher={Oxford University Press}
}

@article{chiang2023vicuna,
  title={Vicuna: An open-source chatbot impressing gpt-4 with 90\%* chatgpt quality},
  author={Chiang, Wei-Lin and Li, Zhuohan and Lin, Ziqing and Sheng, Ying and Wu, Zhanghao and Zhang, Hao and Zheng, Lianmin and Zhuang, Siyuan and Zhuang, Yonghao and Gonzalez, Joseph E and others},
  journal={See https://vicuna. lmsys. org (accessed 14 April 2023)},
  volume={2},
  number={3},
  pages={6},
  year={2023}
}

@article{huang2024good,
  title={How good are low-bit quantized llama3 models? an empirical study},
  author={Huang, Wei and Ma, Xudong and Qin, Haotong and Zheng, Xingyu and Lv, Chengtao and Chen, Hong and Luo, Jie and Qi, Xiaojuan and Liu, Xianglong and Magno, Michele},
  journal={CoRR},
  year={2024}
}

@article{touvron2023llama2,
  title={Llama 2: Open foundation and fine-tuned chat models},
  author={Touvron, Hugo and Martin, Louis and Stone, Kevin and Albert, Peter and Almahairi, Amjad and Babaei, Yasmine and Bashlykov, Nikolay and Batra, Soumya and Bhargava, Prajjwal and Bhosale, Shruti and others},
  journal={arXiv preprint arXiv:2307.09288},
  year={2023}
}

@article{achiam2023gpt,
  title={Gpt-4 technical report},
  author={Achiam, Josh and Adler, Steven and Agarwal, Sandhini and Ahmad, Lama and Akkaya, Ilge and Aleman, Florencia Leoni and Almeida, Diogo and Altenschmidt, Janko and Altman, Sam and Anadkat, Shyamal and others},
  journal={arXiv preprint arXiv:2303.08774},
  year={2023}
}

@article{yang2025qwen3,
  title={Qwen3 technical report},
  author={Yang, An and Li, Anfeng and Yang, Baosong and Zhang, Beichen and Hui, Binyuan and Zheng, Bo and Yu, Bowen and Gao, Chang and Huang, Chengen and Lv, Chenxu and others},
  journal={arXiv preprint arXiv:2505.09388},
  year={2025}
}

@article{liu2023visual,
  title={Visual instruction tuning},
  author={Liu, Haotian and Li, Chunyuan and Wu, Qingyang and Lee, Yong Jae},
  journal={Advances in neural information processing systems},
  volume={36},
  pages={34892--34916},
  year={2023}
}

@article{li2025otter,
  title={Otter: A multi-modal model with in-context instruction tuning},
  author={Li, Bo and Zhang, Yuanhan and Chen, Liangyu and Wang, Jinghao and Pu, Fanyi and Cahyono, Joshua Adrian and Yang, Jingkang and Li, Chunyuan and Liu, Ziwei},
  journal={IEEE Transactions on Pattern Analysis and Machine Intelligence},
  year={2025},
  publisher={IEEE}
}

@article{team2025gemma,
  title={Gemma 3 technical report},
  author={Team, Gemma and Kamath, Aishwarya and Ferret, Johan and Pathak, Shreya and Vieillard, Nino and Merhej, Ramona and Perrin, Sarah and Matejovicova, Tatiana and Ram{\'e}, Alexandre and Rivi{\`e}re, Morgane and others},
  journal={arXiv preprint arXiv:2503.19786},
  year={2025}
}

@article{liu2023mitigating,
  title={Mitigating hallucination in large multi-modal models via robust instruction tuning},
  author={Liu, Fuxiao and Lin, Kevin and Li, Linjie and Wang, Jianfeng and Yacoob, Yaser and Wang, Lijuan},
  journal={arXiv preprint arXiv:2306.14565},
  year={2023}
}

@article{li2025cai,
  title={CAI: Caption-Sensitive Attention Intervention for Mitigating Object Hallucination in Large Vision-Language Models},
  author={Li, Qiming and Ye, Zekai and Feng, Xiaocheng and Zhong, Weihong and Qin, Libo and Chen, Ruihan and Li, Baohang and Jiang, Kui and Wang, Yaowei and Liu, Ting and others},
  journal={arXiv preprint arXiv:2506.23590},
  year={2025}
}

@inproceedings{yu2024hallucidoctor,
  title={Hallucidoctor: Mitigating hallucinatory toxicity in visual instruction data},
  author={Yu, Qifan and Li, Juncheng and Wei, Longhui and Pang, Liang and Ye, Wentao and Qin, Bosheng and Tang, Siliang and Tian, Qi and Zhuang, Yueting},
  booktitle={Proceedings of the IEEE/CVF Conference on Computer Vision and Pattern Recognition},
  pages={12944--12953},
  year={2024}
}

@inproceedings{tang2025seeing,
  title={Seeing Far and Clearly: Mitigating Hallucinations in MLLMs with Attention Causal Decoding},
  author={Tang, Feilong and Liu, Chengzhi and Xu, Zhongxing and Hu, Ming and Huang, Zile and Xue, Haochen and Chen, Ziyang and Peng, Zelin and Yang, Zhiwei and Zhou, Sijin and others},
  booktitle={Proceedings of the Computer Vision and Pattern Recognition Conference},
  pages={26147--26159},
  year={2025}
}

@inproceedings{
zhang2024how,
title={How Language Model Hallucinations Can Snowball},
author={Muru Zhang and Ofir Press and William Merrill and Alisa Liu and Noah A. Smith},
booktitle={Forty-first International Conference on Machine Learning},
year={2024},
url={https://openreview.net/forum?id=FPlaQyAGHu}
}

@article{gan2025textual,
  title={Textual Steering Vectors Can Improve Visual Understanding in Multimodal Large Language Models},
  author={Gan, Woody Haosheng and Fu, Deqing and Asilis, Julian and Liu, Ollie and Yogatama, Dani and Sharan, Vatsal and Jia, Robin and Neiswanger, Willie},
  journal={arXiv preprint arXiv:2505.14071},
  year={2025}
}

@inproceedings{an2025mitigating,
  title={Mitigating object hallucinations in large vision-language models with assembly of global and local attention},
  author={An, Wenbin and Tian, Feng and Leng, Sicong and Nie, Jiahao and Lin, Haonan and Wang, QianYing and Chen, Ping and Zhang, Xiaoqin and Lu, Shijian},
  booktitle={Proceedings of the Computer Vision and Pattern Recognition Conference},
  pages={29915--29926},
  year={2025}
}

@inproceedings{suo2025octopus,
  title={Octopus: Alleviating hallucination via dynamic contrastive decoding},
  author={Suo, Wei and Zhang, Lijun and Sun, Mengyang and Wu, Lin Yuanbo and Wang, Peng and Zhang, Yanning},
  booktitle={Proceedings of the Computer Vision and Pattern Recognition Conference},
  pages={29904--29914},
  year={2025}
}

@inproceedings{wu2025antidote,
  title={Antidote: A Unified Framework for Mitigating LVLM Hallucinations in Counterfactual Presupposition and Object Perception},
  author={Wu, Yuanchen and Zhang, Lu and Yao, Hang and Du, Junlong and Yan, Ke and Ding, Shouhong and Wu, Yunsheng and Li, Xiaoqiang},
  booktitle={Proceedings of the Computer Vision and Pattern Recognition Conference},
  pages={14646--14656},
  year={2025}
}

@inproceedings{yang2025nullu,
  title={Nullu: Mitigating object hallucinations in large vision-language models via halluspace projection},
  author={Yang, Le and Zheng, Ziwei and Chen, Boxu and Zhao, Zhengyu and Lin, Chenhao and Shen, Chao},
  booktitle={Proceedings of the Computer Vision and Pattern Recognition Conference},
  pages={14635--14645},
  year={2025}
}

@inproceedings{yin2025clearsight,
  title={ClearSight: Visual Signal Enhancement for Object Hallucination Mitigation in Multimodal Large Language Models},
  author={Yin, Hao and Si, Guangzong and Wang, Zilei},
  booktitle={Proceedings of the Computer Vision and Pattern Recognition Conference},
  pages={14625--14634},
  year={2025}
}

@inproceedings{chen2024driving,
  title={Driving with llms: Fusing object-level vector modality for explainable autonomous driving},
  author={Chen, Long and Sinavski, Oleg and H{\"u}nermann, Jan and Karnsund, Alice and Willmott, Andrew James and Birch, Danny and Maund, Daniel and Shotton, Jamie},
  booktitle={2024 IEEE International Conference on Robotics and Automation (ICRA)},
  pages={14093--14100},
  year={2024},
  organization={IEEE}
}

@inproceedings{chen2025ict,
  title={Ict: Image-object cross-level trusted intervention for mitigating object hallucination in large vision-language models},
  author={Chen, Junzhe and Zhang, Tianshu and Huang, Shiyu and Niu, Yuwei and Zhang, Linfeng and Wen, Lijie and Hu, Xuming},
  booktitle={Proceedings of the Computer Vision and Pattern Recognition Conference},
  pages={4209--4221},
  year={2025}
}

@article{wang2024mitigating,
  title={Mitigating hallucinations in large vision-language models with instruction contrastive decoding},
  author={Wang, Xintong and Pan, Jingheng and Ding, Liang and Biemann, Chris},
  journal={arXiv preprint arXiv:2403.18715},
  year={2024}
}

@article{li2023inference,
  title={Inference-time intervention: Eliciting truthful answers from a language model},
  author={Li, Kenneth and Patel, Oam and Vi{\'e}gas, Fernanda and Pfister, Hanspeter and Wattenberg, Martin},
  journal={Advances in Neural Information Processing Systems},
  volume={36},
  pages={41451--41530},
  year={2023}
}

@article{abdi2010principal,
  title={Principal component analysis},
  author={Abdi, Herv{\'e} and Williams, Lynne J},
  journal={Wiley interdisciplinary reviews: computational statistics},
  volume={2},
  number={4},
  pages={433--459},
  year={2010},
  publisher={Wiley Online Library}
}

@book{vasiliev2020natural,
  title={Natural language processing with Python and spaCy: A practical introduction},
  author={Vasiliev, Yuli},
  year={2020},
  publisher={No Starch Press}
}

@article{wan2024look,
  title={Look-m: Look-once optimization in kv cache for efficient multimodal long-context inference},
  author={Wan, Zhongwei and Wu, Ziang and Liu, Che and Huang, Jinfa and Zhu, Zhihong and Jin, Peng and Wang, Longyue and Yuan, Li},
  journal={arXiv preprint arXiv:2406.18139},
  year={2024}
}

@article{tu2024vl,
  title={VL-cache: Sparsity and modality-aware KV cache compression for vision-language model inference acceleration},
  author={Tu, Dezhan and Vashchilenko, Danylo and Lu, Yuzhe and Xu, Panpan},
  journal={arXiv preprint arXiv:2410.23317},
  year={2024}
}

@inproceedings{xu2024ote,
  title={Ote: Exploring accurate scene text recognition using one token},
  author={Xu, Jianjun and Wang, Yuxin and Xie, Hongtao and Zhang, Yongdong},
  booktitle={Proceedings of the IEEE/CVF Conference on Computer Vision and Pattern Recognition},
  pages={28327--28336},
  year={2024}
}

@article{wan2025meda,
  title={Meda: Dynamic kv cache allocation for efficient multimodal long-context inference},
  author={Wan, Zhongwei and Shen, Hui and Wang, Xin and Liu, Che and Mai, Zheda and Zhang, Mi},
  journal={arXiv preprint arXiv:2502.17599},
  year={2025}
}

@article{wang2024model,
  title={Model tells you where to merge: Adaptive kv cache merging for llms on long-context tasks},
  author={Wang, Zheng and Jin, Boxiao and Yu, Zhongzhi and Zhang, Minjia},
  journal={arXiv preprint arXiv:2407.08454},
  year={2024}
}

@inproceedings{wang-etal-2025-metok,
    title = "{MET}ok: Multi-Stage Event-based Token Compression for Efficient Long Video Understanding",
    author = "Wang, Mengyue  and
      Chen, Shuo  and
      Kersting, Kristian  and
      Tresp, Volker  and
      Ma, Yunpu",
    editor = "Christodoulopoulos, Christos  and
      Chakraborty, Tanmoy  and
      Rose, Carolyn  and
      Peng, Violet",
    booktitle = "Proceedings of the 2025 Conference on Empirical Methods in Natural Language Processing",
    month = nov,
    year = "2025",
    address = "Suzhou, China",
    publisher = "Association for Computational Linguistics",
    url = "https://aclanthology.org/2025.emnlp-main.954/",
    doi = "10.18653/v1/2025.emnlp-main.954",
    pages = "18881--18895",
    ISBN = "979-8-89176-332-6", 
}

@inproceedings{li-etal-2025-fairsteer,
    title = "{F}air{S}teer: Inference Time Debiasing for {LLM}s with Dynamic Activation Steering",
    author = "Li, Yichen  and
      Fan, Zhiting  and
      Chen, Ruizhe  and
      Gai, Xiaotang  and
      Gong, Luqi  and
      Zhang, Yan  and
      Liu, Zuozhu",
    editor = "Che, Wanxiang  and
      Nabende, Joyce  and
      Shutova, Ekaterina  and
      Pilehvar, Mohammad Taher",
    booktitle = "Findings of the Association for Computational Linguistics: ACL 2025",
    month = jul,
    year = "2025",
    address = "Vienna, Austria",
    publisher = "Association for Computational Linguistics",
    url = "https://aclanthology.org/2025.findings-acl.589/",
    doi = "10.18653/v1/2025.findings-acl.589",
    pages = "11293--11312",
    ISBN = "979-8-89176-256-5"
}

@article{belitsky2025kv,
  title={Kv cache steering for inducing reasoning in small language models},
  author={Belitsky, Max and Kopiczko, Dawid J and Dorkenwald, Michael and Jehanzeb Mirza, M and Snoek, Cees GM and Asano, Yuki M},
  journal={arXiv e-prints},
  pages={arXiv--2507},
  year={2025}
}

@inproceedings{yang2025improving,
  title={Improving factuality in large language models via decoding-time hallucinatory and truthful comparators},
  author={Yang, Dingkang and Xiao, Dongling and Wei, Jinjie and Li, Mingcheng and Chen, Zhaoyu and Li, Ke and Zhang, Lihua},
  booktitle={Proceedings of the AAAI Conference on Artificial Intelligence},
  volume={39},
  number={24},
  pages={25606--25614},
  year={2025}
}

@inproceedings{
fu2025not,
title={Not All Heads Matter: A Head-Level {KV} Cache Compression Method with Integrated Retrieval and Reasoning},
author={Yu Fu and Zefan Cai and Abedelkadir Asi and Wayne Xiong and Yue Dong and Wen Xiao},
booktitle={The Thirteenth International Conference on Learning Representations},
year={2025},
url={https://openreview.net/forum?id=FJFVmeXusW}
}

@article{sridhar2025video,
  title={Video Reasoning without Training},
  author={Sridhar, Deepak and Bhardwaj, Kartikeya and Jeyaraj, Jeya Pradha and Vasconcelos, Nuno and Nayak, Ankita and Teague, Harris},
  journal={arXiv preprint arXiv:2510.17045},
  year={2025}
}

@article{wu2025fast,
  title={Fast-dllm: Training-free acceleration of diffusion llm by enabling kv cache and parallel decoding},
  author={Wu, Chengyue and Zhang, Hao and Xue, Shuchen and Liu, Zhijian and Diao, Shizhe and Zhu, Ligeng and Luo, Ping and Han, Song and Xie, Enze},
  journal={arXiv preprint arXiv:2505.22618},
  year={2025}
}

@article{vaswani2017attention,
  title={Attention is all you need},
  author={Vaswani, Ashish and Shazeer, Noam and Parmar, Niki and Uszkoreit, Jakob and Jones, Llion and Gomez, Aidan N and Kaiser, {\L}ukasz and Polosukhin, Illia},
  journal={Advances in neural information processing systems},
  volume={30},
  year={2017}
}

@inproceedings{bi2025llava,
  title={Llava steering: Visual instruction tuning with 500x fewer parameters through modality linear representation-steering},
  author={Bi, Jinhe and Wang, Yujun and Chen, Haokun and Xiao, Xun and Hecker, Artur and Tresp, Volker and Ma, Yunpu},
  booktitle={Proceedings of the 63rd Annual Meeting of the Association for Computational Linguistics (Volume 1: Long Papers)},
  pages={15230--15250},
  year={2025}
}

@article{vogels2025distribution,
  title={In-Distribution Steering: Balancing Control and Coherence in Language Model Generation},
  author={Vogels, Arthur and Wong, Benjamin and Choho, Yann and Blangero, Annabelle and Bhan, Milan},
  journal={arXiv preprint arXiv:2510.13285},
  year={2025}
}

@article{liu2024deliberation,
  title={Deliberation in latent space via differentiable cache augmentation},
  author={Liu, Luyang and Pfeiffer, Jonas and Wu, Jiaxing and Xie, Jun and Szlam, Arthur},
  journal={arXiv preprint arXiv:2412.17747},
  year={2024}
}

@article{hurst2024gpt,
  title={Gpt-4o system card},
  author={Hurst, Aaron and Lerer, Adam and Goucher, Adam P and Perelman, Adam and Ramesh, Aditya and Clark, Aidan and Ostrow, AJ and Welihinda, Akila and Hayes, Alan and Radford, Alec and others},
  journal={arXiv preprint arXiv:2410.21276},
  year={2024}
}
}

\maketitlesupplementary

\appendix
\section{Details of Internal Interpretability Analysis.}
\label{appendix:interp}
In this section, we provide the mathematical formulation and calculation steps of the change rate of attention for the internal interpretability analysis presented in Section \ref{analysis:Internal}. 

Specifically, we randomly sample $300$ images from the MSCOCO \cite{lin2014microsoft} dataset. 
For each image, we denote $\mathbf{A}^{t}_{l,h} \in \mathbb{R}^{N}$ as the last token attention distribution of the $h$-th head in the $l$-th layer at generation step $t$ over the sequence of length $N$, 
$\mathcal{I}_{\text{img}}$ represent the set of indices corresponding to visual tokens. 
Then, we quantitatively measure the rate of change in both 1) Global Visual Attention and 2) Local Object-Centric Attention before and after our intervention.

\subsection{Global Visual Attention Dynamics}
To quantify how the model's reliance on visual information evolves throughout the generation process, we employ a stage-wise attention analysis. This allows us to monitor the ``attention decay'' phenomenon \cite{tang2025seeing,liu2024paying,li2025cai} and verify whether our intervention effectively counteracts this trend.

\noindent\textbf{Visual Attention Proportion.} 
First, for any given generation step $t$, we define the global visual attention proportion $P_{\text{img}}^{t}$ as follows:
\begin{equation}
P_{\text{img}}^{t} = \frac{1}{L \cdot H} \sum_{l=1}^{L} \sum_{h=1}^{H} \frac{\sum_{v \in \mathcal{I}_{\text{img}}} \mathbf{A}^{t}_{l,h}[v]}{\sum_{n\in N} \mathbf{A}^{t}_{l,h}[n] + \epsilon}
\end{equation}
which represents the average probability mass allocated to visual tokens across all layers $L$ and heads $H$. Here, $\epsilon$ is a small constant for numerical stability.

\noindent\textbf{Temporal Alignment via Staged Sampling.} 
Since the total length of generated tokens $N$ varies between the original (before) and intervened (after) models, we normalize the generation process into discrete progression stages to ensure a fair alignment. 
Specifically, we divide the generation into $K$ discrete stages.
For the $k$-th stage, we sample the proportion of visual attention at step $t_k$ as follows:
\begin{equation}
t_k= \lfloor \frac{k}{K} \cdot N \rfloor, \quad k \in \{0, 1, \dots, K\}
\end{equation}
which effectively captures the behavior of the model from the onset ($0\%$) to the completion ($100\%$) of the response.

\noindent\textbf{Global Relative Change Rate.} 
To evaluate the impact of our intervention, we calculate the relative change in visual attention at each aligned stage. Let $P_{\text{img}}^{t_k}$ and ${\hat P}_{\text{img}}^{t_k}$ denote the visual attention proportions for the vanilla (before) and intervened (after) models at stage $k$, respectively. The stage-wise relative change rate $\Delta R^{k}$ is defined as follows:
\begin{equation}
\Delta R^{k} = \frac{{\hat P}_{\text{img}}^{t_k} - P_{\text{img}}^{t_k}}{ \left| P_{\text{img}}^{t_k} \right| + \epsilon} \times 100\%.
\end{equation}
Here, a positive $\Delta R^{k}$ indicates that the intervention effectively enhances the model's visual grounding capabilities relative to the Vanilla.
Crucially, by analyzing the trend of $\Delta R =\{\Delta R^0, \Delta R^1, \dots, \Delta R^K\}$, we can verify whether the intervention specifically counteracts the attention decay that typically occurs in the later stages of long-form generation.

\subsection{Local Object-Centric Attention Shift}
Beyond preserving the global magnitude of visual signals, we further verify whether the enhanced visual attention is meaningfully focused on the relevant object regions rather than the background.

\noindent\textbf{Visual Attention Re-normalization.} 
We first disentangle the spatial distribution from the total visual attention weight.
For a specific layer $l$ and head $h$ at step $t$, we compute the conditional probability distribution $\tilde{\mathbf{A}}^{t}_{l,h}$ over the visual tokens:
\begin{equation}
\tilde{\mathbf{A}}^{t}_{l,h}[v] = \frac{\mathbf{A}^{t}_{l,h}[v]}{\sum_{j \in \mathcal{I}_{\text{img}}} \mathbf{A}^{t}_{l,h}[j]}, \quad \forall v \in \mathcal{I}_{\text{img}}
\end{equation}
This re-normalization reflects the spatial focus of the head, \textit{given} that it is attending to the image.

\noindent\textbf{Object-Centric Attention Score.} 
Let $M \in \{0, 1\}^{|\mathcal{I}_{\text{img}}|}$ be the binary mask of the ground-truth object, where $M[v]=1$ indicates token $v$ falls within the object region.
We calculate the aggregated attention mass allocated to the object as:
\begin{equation}
S_{\text{obj}}^{t}(l, h) = \sum_{v \in \mathcal{I}_{\text{img}}} \tilde{\mathbf{A}}^{t}_{l,h}[v] \cdot M[v]
\end{equation}
This metric ($S_{\text{obj}}$) provides a granular view of the model's spatial grounding capability at the head level.

\begin{table*}[!th]
    \vspace{-5pt}
    \centering
    \caption{
    More results on the random and popular splits of the POPE benchmark. 
    }
      \vspace{-5pt}
    \label{tab:POPE_RP}
    \resizebox{1\linewidth}{!}{
    \begin{tabular}{lcccccccccccc}
        \whline
        \multirow{3}{*}{Method } 
        &\multicolumn{4}{c}{LLAVA-1.5} & \multicolumn{4}{c}{Qwen-VL-Chat} & \multicolumn{4}{c}{DeepSeek-VL-Chat}\\
        \cmidrule(r){2-5} \cmidrule(lr){6-9} \cmidrule(l){10-13} 
        & \multicolumn{2}{c}{Random} & \multicolumn{2}{c}{Popular} 
        & \multicolumn{2}{c}{Random} & \multicolumn{2}{c}{Popular}  
        & \multicolumn{2}{c}{Random} & \multicolumn{2}{c}{Popular}  \\
         
        \cmidrule(r){2-3} \cmidrule(lr){4-5} \cmidrule(lr){6-7} \cmidrule(lr){8-9} \cmidrule(lr){10-11} \cmidrule(l){12-13} 
 
         ~ &   $\text{Acc} \uparrow$  & $\text{F1} \uparrow$  &  $\text{Acc} \uparrow$  & $\text{F1} \uparrow$  &  $\text{Acc} \uparrow$  & $\text{F1} \uparrow$   &  $\text{Acc} \uparrow$  & $\text{F1} \uparrow$ &  $\text{Acc} \uparrow$  & $\text{F1} \uparrow$ &  $\text{Acc} \uparrow$  & $\text{F1} \uparrow$  \\

        \hline

\cellcolor{mygray}Vanilla  & \cellcolor{mygray}$83.74$ & \cellcolor{mygray}$84.54$ &  \cellcolor{mygray}$80.50$ & \cellcolor{mygray}$81.49$ 
& \cellcolor{mygray}$86.52$ & \cellcolor{mygray}$85.59$ & \cellcolor{mygray}$84.30$ & \cellcolor{mygray}$83.39$  
& \cellcolor{mygray}$84.81$ & \cellcolor{mygray}$84.18$ & \cellcolor{mygray}84.13  & \cellcolor{mygray}$83.24$  \\

PAI \cite{liu2024paying}  &  $\bold{85.73}$ & $\bold{86.32}$ & $82.53$ & $83.40$ 
& $\bold{87.69}$ & $\bold{86.94}$ & \underline{$85.40$} & \underline{$84.40$}
& $\bold{86.83}$ & $\bold{86.29}$ & \underline{$85.86$} & \underline{$84.93$} \\

 VCD \cite{leng2024mitigating} &$85.36$ & \underline{$86.00$} & \underline{$82.60$} & \underline{$83.41$} 
& $86.93$  & $85.46$  & $85.17$  &  $83.68$ 
&  $86.70$  & $85.97$  & $85.80$  & $84.56$  \\

 VTI \cite{liu2024reducing}  & $83.40$  &  $84.14$ & $81.50$ & $82.33$ 
& $85.77$ & $84.68$ & $83.86$ & $82.59$ 
&  $86.15$  &  $86.01$ & $84.83$  & $84.48$  \\

VISTA \cite{li2025hidden} & $84.74$ & $85.78$ & $79.97$ & $81.55$ 
& $85.91$ & $84.66$ & $83.90$ & $82.25$  
&  \underline{$86.70$} & $85.54$ & $\bold{86.13}$ & $84.58$  \\

\cellcolor{myblue}\textbf{PTI (ours)} &  
\cellcolor{myblue}\underline{$85.43$} &  
\cellcolor{myblue}$85.72$ &  
\cellcolor{myblue}$\bold{83.80}$ &  
\cellcolor{myblue}$\bold{84.09}$ &
\cellcolor{myblue}\underline{$87.53$} &  
\cellcolor{myblue}\underline{$86.54$} & 
\cellcolor{myblue}$\bold{86.17}$ &  
\cellcolor{myblue}$\bold{84.89}$ &
\cellcolor{myblue}$86.22$  &  
\cellcolor{myblue}\underline{$86.25$} &  
\cellcolor{myblue}$85.43$ &  
\cellcolor{myblue}$\bold{85.16}$  \\ 
   
        \whline
        \end{tabular}
    }
\end{table*}
\begin{table*}[!ht]
    \centering
    \caption{Evaluation of MME Benchmark. ``Exist.'': Existence. ``Posit.'': Position. The maximum new token is set to 32.
    }
    \vspace{-5pt}
    \resizebox{1\linewidth}{!}{%
    \begin{tabular}{lcccc|c|cccc|c|cccc|c}
        \whline
\multirow{2}{*}{Method}           & \multicolumn{5}{c}{LLAVA-1.5}  & \multicolumn{5}{c}{Qwen-VL-Chat}   & \multicolumn{5}{c}{DeepSeek-VL-Chat} \\
        \cmidrule(r){2-6} \cmidrule(r){7-11}  \cmidrule(r){12-16} 

                &  Exist. & Count & Posit. & Color & \textbf{Total}   &  Exist. & Count & Posit. & Color &  \textbf{Total}   &  Exist. & Count & Posit. & Color & \textbf{Total} \\
\hline
\rowcolor{mygray}
Vanila &  \tabincell{c}{$180.0$} & \tabincell{c}{$143.3$} & \tabincell{c}{${133.3}$} & \tabincell{c}{$155.0$} & \tabincell{c}{$611.6$} 
&  \tabincell{c}{$180.0$} & \tabincell{c}{$115.0$} & \tabincell{c}{$128.3$} & \tabincell{c}{$175.0$} & \tabincell{c}{$598.3$} 
&  \tabincell{c}{$190.0$} & \tabincell{c}{$153.3$} & \tabincell{c}{$133.3$} & \tabincell{c}{$175.0$} & \tabincell{c}{$651.6$} 
\\

PAI \cite{liu2024paying} &  \tabincell{c}{$190.0$} & \tabincell{c}{$148.3$} & \tabincell{c}{$126.6$} & \tabincell{c}{$160.0$} & \tabincell{c}{$625.0$} 
&  \tabincell{c}{$180.0$} & \tabincell{c}{$120.0$} & \tabincell{c}{$130.0$} & \tabincell{c}{$175.0$} & \tabincell{c}{$605.0$}
&  \tabincell{c}{$190.0$} & \tabincell{c}{$158.3$} & \tabincell{c}{$133.3$} & \tabincell{c}{$175.0$} & \tabincell{c}{$656.6$} 
\\

VTI \cite{liu2024reducing} &  \tabincell{c}{$185.0$} & \tabincell{c}{$153.3$} & \tabincell{c}{{${130.0}$}} & \tabincell{c}{$165.0$} & \tabincell{c}{$\underline{633.3}$} 
&  \tabincell{c}{$185.0$} & \tabincell{c}{\textbf{${145.0}$}} & \tabincell{c}{$116.6$} & \tabincell{c}
{\textbf{${180.0}$}} & \tabincell{c}{$\underline{626.6}$}
&  \tabincell{c}{$195.0$} & \tabincell{c}{$158.3$} & \tabincell{c}{$133.3$} & \tabincell{c}{$175.0$} & \tabincell{c}{$\underline{661.6}$} 
\\

VISTA  \cite{li2025hidden} &  \tabincell{c}{$195.0$} & \tabincell{c}{$138.3$} & \tabincell{c}{$121.6$} & \tabincell{c}{$160.0$} & \tabincell{c}{$615.0$} 
&  \tabincell{c}{$175.0$} & \tabincell{c}{$125.0$} & \tabincell{c}{$141.6$} & \tabincell{c}{$170.0$} & \tabincell{c}{$611.6$}
&  \tabincell{c}{$185.0$} & \tabincell{c}{$158.3$} & \tabincell{c}{$128.3$} & \tabincell{c}{$175.0$} & \tabincell{c}{$646.6$} 

\\

\rowcolor{myblue}
\textbf{PTI (ours)} &  \tabincell{c}{\textbf{${195.0}$}} & \tabincell{c}{{\textbf{${163.3}$}}} & \tabincell{c}{$128.3$} & \tabincell{c}{\textbf{${165.0}$}} & \tabincell{c}{\textbf{$\bold{651.6}$}} 
&  \tabincell{c}{\textbf{${185.0}$}} & \tabincell{c}{{$140.0$}} & \tabincell{c}{\textbf{${148.3}$}} & \tabincell{c}{{${165.0}$}} & \tabincell{c}{\textbf{$\bold{638.3}$}} 
&  \tabincell{c}{\textbf{${195.0}$}} & \tabincell{c}{\textbf{${163.3}$}} & \tabincell{c}{\textbf{${138.3}$}} & \tabincell{c}{\textbf{$175.0$}} & \tabincell{c}{\textbf{$\bold{671.6}$}}
\\
        \whline
    \end{tabular}}
    \label{tab:mme_full}
\end{table*}

\noindent\textbf{Local Targeted Shift}
Finally, to visualize the layer-wise and head-wise impact of our intervention, we compute the absolute shift in object-centric attention:
Let $S_{\text{obj}}^{t}(l, h)$ and $\hat S_{\text{obj}}^{t}(l, h)$ denote the proportions of object attention for the vanilla (before) and intervened (after) models at step $t$, respectively. The object-centric attention shift $\Delta S^t_{\text{obj}}(l, h)$ is defined as follows:
\begin{equation}
\Delta S^t_{\text{obj}}(l, h) = \hat S_{\text{obj}}^{t}(l, h) - S_{\text{obj}}^{t}(l, h).
\end{equation}
A positive $\Delta S_{\text{obj}}(l, h)$ signifies that the $h$-th head in the $l$-th layer has successfully redistributed probability mass from the background to the object region. 
We visualize these shifts as heatmaps (see Figure \ref{analysis:total} Right) to identify which specific components of the Transformer are responsible for the improved object grounding.

\section{Additional Experiments of PTI.}
\label{appendix:exper}

\subsection{More Experimental Results of POPE.}

Table \ref{tab:POPE_RP} presents the detailed performance on the \textit{Random} and \textit{Popular} splits of the POPE benchmark, supplementing the Adversarial results discussed in Section \ref{exp:pope}.
Consistent with the performance in Table \ref{tab:POPE}, PTI demonstrates robust generalization capabilities across varying difficulty levels.
Specifically, in the \textit{Popular} split, which challenges models with high-frequency objects prone to statistical language priors, PTI achieves superior performance, surpassing the Vanilla baseline by substantial margins (e.g., $+2.60\%$ F1 on LLaVA-1.5 and $+1.50\%$ F1 on Qwen-VL-Chat) and outperforming competitive baselines like PAI \cite{liu2024paying} and VCD \cite{leng2024mitigating}.
In the \textit{Random} split, where performance is generally saturated due to lower difficulty, PTI maintains competitive results, consistently ranking within the top-two across all evaluated LVLMs.
Notably, PTI consistently outperforms both VTI \cite{liu2024reducing} and VISTA \cite{li2025hidden}, offering a significantly more effective steering paradigm.

\subsection{Detailed Experimental Results of MME.}
Table \ref{tab:mme_full} details the complete performance on the MME benchmark for evaluating object-level and attribute-level hallucination.
Notably, PTI exhibits substantial gains in the most challenging fine-grained perception tasks, specifically ``Count'' and ``Position'' where standard models frequently struggle. 
While the competitive VTI method shows strength in specific attributes like color, PTI maintains a superior balance across all dimensions. 
This consistent superiority in spatially sensitive and quantitative metrics validates that our PTI effectively sharpens the model's initial visual grounding, enabling it to better resolve distinct objects and their spatial relationships before the decoding phase begins.

\begin{table*}[h]
    \centering
    \caption{Measure of Latency (ms/token) and Throughput (token/s) on CHAIR benchmark. All results use the Nucleus Sampling decoding strategy on a NVIDIA 4090 GPU.}
    \vspace{-5pt}
    \resizebox{1\linewidth}{!}{%
    \begin{tabular}{lccccccc}
        \whline
        \multirow{2}{*}{Method}           
        & \multicolumn{2}{c}{LLAVA-1.5} 
        & \multicolumn{2}{c}{Qwen-VL-Chat} 
        & \multicolumn{2}{c}{DeepSeek-VL-Chat} \\
        
        \cmidrule(r){2-3} \cmidrule(r){4-5}  \cmidrule(r){6-7}  

         & $\text{Latency   } \downarrow$ & $\text{Throughput } \uparrow $ 
         & $\text{Latency } \downarrow$ & $\text{Throughput } \uparrow $ 
         & $\text{Latency }  \downarrow$ & $\text{Throughput } \uparrow $  \\


\hline
\rowcolor{mygray}
Vanilla  
&\tabincell{c}{$19.52$ ($\times1.00)$}   
&\tabincell{c}{$51.22$ ($\times 1.00$)}
& \tabincell{c}{$20.55$  ($\times1.00$)} 
&\tabincell{c}{$48.66$ ($\times1.00$)}
& \tabincell{c}{$18.29$  ($\times1.00$)}
& \tabincell{c}{$54.67$  ($\times1.00$)}
\\

VCD  \cite{leng2024mitigating} 
&\tabincell{c}{$38.92$ \color{mydarkgray}($\times1.99$)}  
& \tabincell{c}{$25.69$ \color{mydarkgray}($\times0.50$)}
&  \tabincell{c}{$41.08$ \color{mydarkgray}($\times1.99$)}
&\tabincell{c}{$24.34$ \color{mydarkgray}($\times0.50$)}
& \tabincell{c}{$36.38$ \color{mydarkgray}($\times1.98$)}
&\tabincell{c}{$27.49$ \color{mydarkgray}($\times0.50$)}
\\

PAI \cite{liu2024paying} 
& \tabincell{c}{$37.62$ \color{mydarkgray}($\times1.93$)}  
&\tabincell{c}{$26.58$ \color{mydarkgray}($\times0.52$)}
& \tabincell{c}{$42.07$ \color{mydarkgray}($\times2.04$)} 
& \tabincell{c}{$23.77$ \color{mydarkgray}($\times0.48$)}
& \tabincell{c}{$35.16$ \color{mydarkgray}($\times1.92$)} 
&\tabincell{c}{$28.44$ \color{mydarkgray}($\times0.52$)} \\

VISTA \cite{li2025hidden} 
&\tabincell{c}{$26.20$ \color{mydarkgray}($\times1.34$)}   
&\tabincell{c}{$38.17$ \color{mydarkgray}($\times0.74$)}
&\tabincell{c}{$33.81$ \color{mydarkgray}($\times1.64$)} 
&\tabincell{c}{$29.57$ \color{mydarkgray}($\times0.60$)}
& \tabincell{c}{$26.79$ \color{mydarkgray}($\times1.46$)}
&\tabincell{c}{$37.33$ \color{mydarkgray}($\times0.68$)} \\

VTI \cite{liu2024reducing} 
& \tabincell{c}{$20.84$ \color{mydarkgray}($\times1.07$)}
&\tabincell{c}{$47.97$ \color{mydarkgray}($\times0.93$)}
& \tabincell{c}{$21.86$ \color{mydarkgray}($\times1.06$)} 
&\tabincell{c}{$45.75$ \color{mydarkgray}($\times0.94$)}
& \tabincell{c}{$19.81$ \color{mydarkgray}($\times1.08$)}  
&\tabincell{c}{$50.47$ \color{mydarkgray}($\times0.92$)} \\

\rowcolor{myblue}
\textbf{PTI (ours) }
& \tabincell{c}{$\bold{19.58}$ \color{mygreen}($\times \bold{1.00}$)} 
& \tabincell{c}{$\bold{51.06}$ \color{mygreen}($\times \bold{0.99}$)} 
& \tabincell{c}{$\bold{20.88}$ \color{mygreen}($\times \bold{1.01}$)} 
& \tabincell{c}{$\bold{47.90}$ \color{mygreen}($\times \bold{0.98}$)}
& \tabincell{c}{$\bold{18.56}$ \color{mygreen}($\times \bold{1.02}$)}  
& \tabincell{c}{$\bold{53.89 }$ \color{mygreen}($\times \bold{0.99}$)}  \\

        \whline
    \end{tabular}}
	\vspace{-5pt}
    \label{tab:latency}
\end{table*}

\begin{figure*}[!th] 
    \centering 
    \begin{minipage}[t]{\linewidth} 
        \centering
        \includegraphics[width=\linewidth]{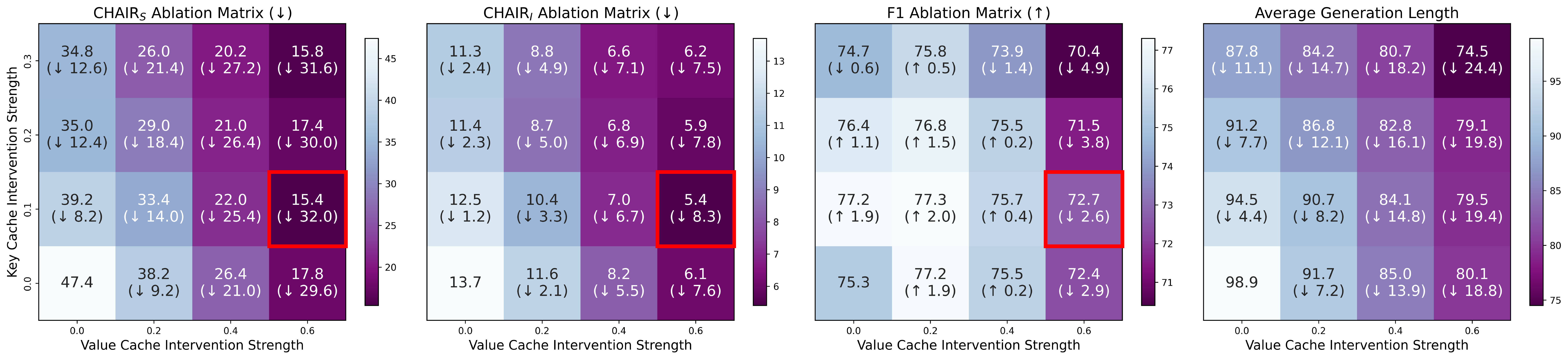} 
    \end{minipage}
    \caption{Ablation matrices for multi-modal KV cache intervention strength on LLAVA-1.5 with greedy decoding strategy. Brighter colors indicate better performance, while red boxes highlight the parameter combinations used in Table \ref{tab:CHAIR}.}
    \label{tables:analysis_ablation_llava}
    

    \begin{minipage}[t]{\linewidth}  
        \centering
        \includegraphics[width=\linewidth]{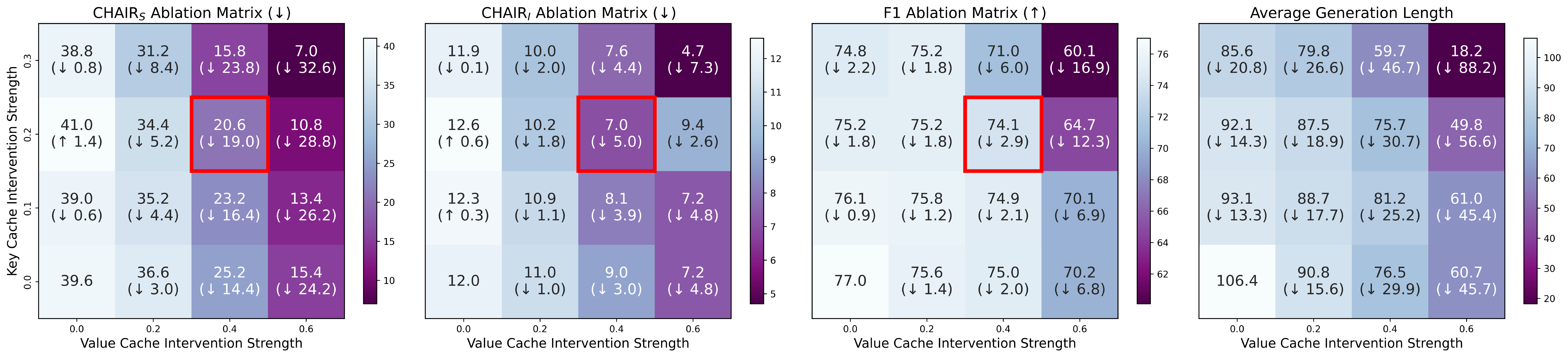} 
    \end{minipage} 
    \caption{Ablation matrices for multi-modal KV cache intervention strength on Qwen-VL-Chat with greedy decoding strategy.}
    \label{tables:analysis_ablation_qwen}


    \begin{minipage}[t]{\linewidth}  
        \centering
        \includegraphics[width=\linewidth]{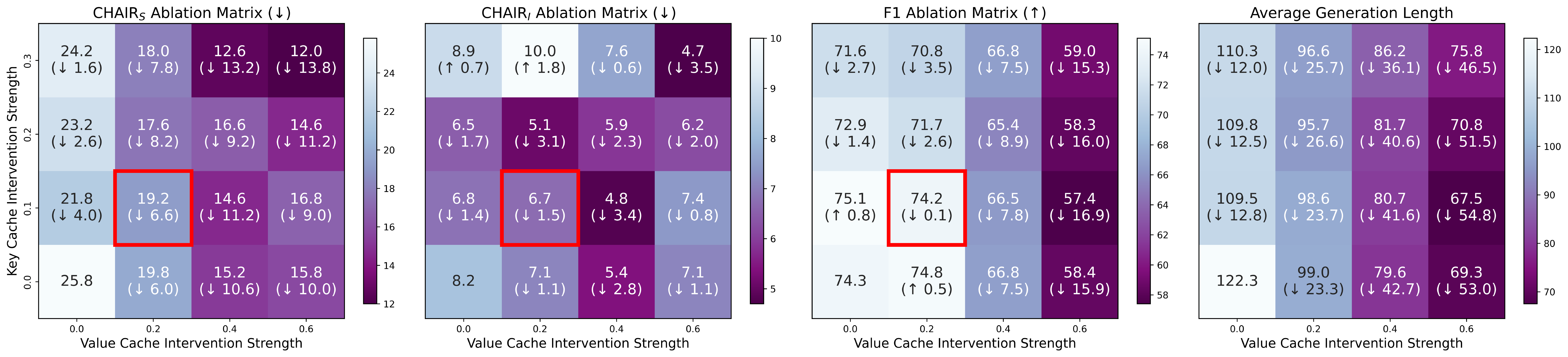} 
    \end{minipage} 
    \caption{Ablation matrices for multi-modal KV cache intervention strength on DeepSeek-VL-Chat with greedy decoding strategy.}
    \label{tables:analysis_ablation_deepseek}

    \vspace{-3pt}
    
\end{figure*}

\subsection{Inference Efficiency Analysis.}
\label{appendix:efficiency}

As shown in Table \ref{tab:latency}, we evaluate the inference efficiency of our method against competing approaches in terms of both latency and throughput. 
Existing methods often introduce significant computational overhead. 
Methods apply case-specific operations, such as VCD \cite{leng2024mitigating} and PAI \cite{liu2024paying}, introduce significant overhead due to contrastive computations, nearly doubling the latency (\textit{e.g.}, $\times1.99$ for VCD on LLaVA-1.5). 
Consequently, their throughput suffers a substantial degradation, dropping to approximately $0.50\times$ of the performance of vanilla.
Additionally, DTI methods like VTI \cite{liu2024reducing}, which apply steering vectors across multiple steps, also incur notable costs (\textit{e.g.}, $\times1.07$, $\times1.06$, and $\times1.08$, respectively), resulting in a throughput decline lower than $\times0.95$.
In contrast, PTI eliminates the need for sample-specific operations or multi-step interventions by modifying the initial KV cache only once. 
As a result, it incurs negligible latency overhead (lower than $\times1.02$ across all models), and sustains near-lossless throughput (over $\times0.98$ across all models). 
This consistent superiority highlights that PTI is a highly efficient, plug-and-play solution suitable for delay-sensitive real-world applications.

\begin{figure*}[!ht] 
    \centering

    \begin{minipage}[t]{\linewidth}
        \centering
        \includegraphics[width=\linewidth]{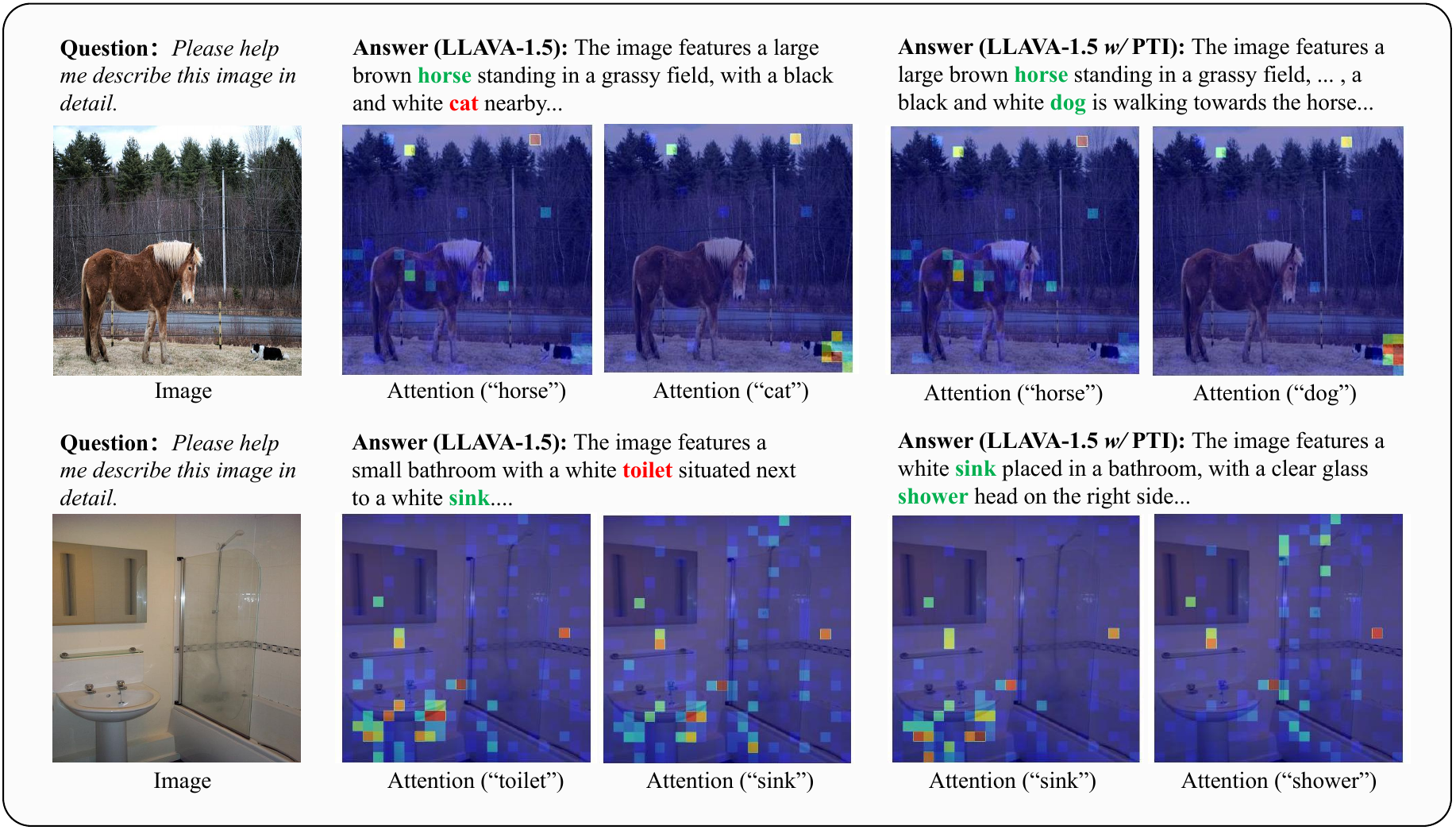} 
    \end{minipage}
    \caption{
    Visual analysis of cross-modal attention maps on LLAVA-1.5. For each sample, the hallucinated content is highlighted in \textcolor{mydarkred}{\textbf{red}}, while the correct content is highlighted in \textcolor{lgreen}{\textbf{green}}.
    The attention map of the target token represents the magnitude of attention weights assigned to image tokens, averaged across all layers and heads.
    }
    \label{tables:analysis_vis_attn} 
\end{figure*}

\subsection{Ablation Experiments of Hyperparameters.}
To investigate the sensitivity of PTI to intervention intensity, we present the ablation matrices for key and value intervention strengths on CHAIR \cite{rohrbach2018object} benchmark across LLaVA-1.5 (Figure \ref{tables:analysis_ablation_llava}), Qwen-VL-Chat (Figure \ref{tables:analysis_ablation_qwen}), and DeepSeek-VL-Chat (Figure \ref{tables:analysis_ablation_deepseek}). 
As mentioned in Section \ref{exp:Implementation}, we unify the visual and textual coefficients by setting $\lambda_{ \text{k,img}}$ = $\lambda_{ \text{k,txt}}$ and $\lambda_{ \text{v,img}}$ = $\lambda_{ \text{v,txt}}$ to reduce the hyperparameter search space.
A consistent trend is observable across these architecture-distinct models: the value cache intervention exerts a more dominant influence on hallucination mitigation, as evidenced by the substantial reduction in $\text{CHAIR}_S$/$\text{CHAIR}_I$ metrics corresponding to increases in value intervention strength. However, the matrices also reveal a critical sensitivity to excessive steering; simultaneously maximizing both key and value intervention strengths precipitates a noticeable degradation in F1 scores and a sharp decline in average generation length, indicating a compromise in the model's generation quality. Consequently, the configurations highlighted in red boxes represent an optimal equilibrium, effectively minimizing hallucination rates via robust visual grounding while preserving the fidelity and completeness of the textual response.

\section{Additional Case Studies}
\label{analysis:case_study}

\subsection{Attention Map Visualization.}
To intuitively understand how PTI mitigates hallucinations, we visualize the cross-modal attention maps of the generated tokens with respect to the visual features. 
As illustrated in Figure \ref{tables:analysis_vis_attn}, PTI substantially enhances the object-centric attention and robust visual recognition of LLAVA-1.5.
Taking the first example as prominent cases, the vanilla model suffers from severe perceptual misalignment and erroneously hallucinates a ``cat''. In contrast, PTI effectively corrects this distribution. 
On one hand, PTI intensifies the attentional weights on the correctly identified dominant object (``horse''), successfully rectifying the model's gaze.
On the other hand, PTI ensures a more robust and stable visual representation, enabling the model to identify the ``dog'' correctly.
This dual effect—strengthening valid signals while correcting misaligned features—confirms that PTI operates by enforcing precise object-centric attention, thereby eliminating object-level hallucinations at their source.

\subsection{Qualitative Examples.}

\begin{figure*}[htbp] 
    \centering

    \begin{minipage}[t]{\linewidth}
        \centering
        \includegraphics[width=\linewidth]{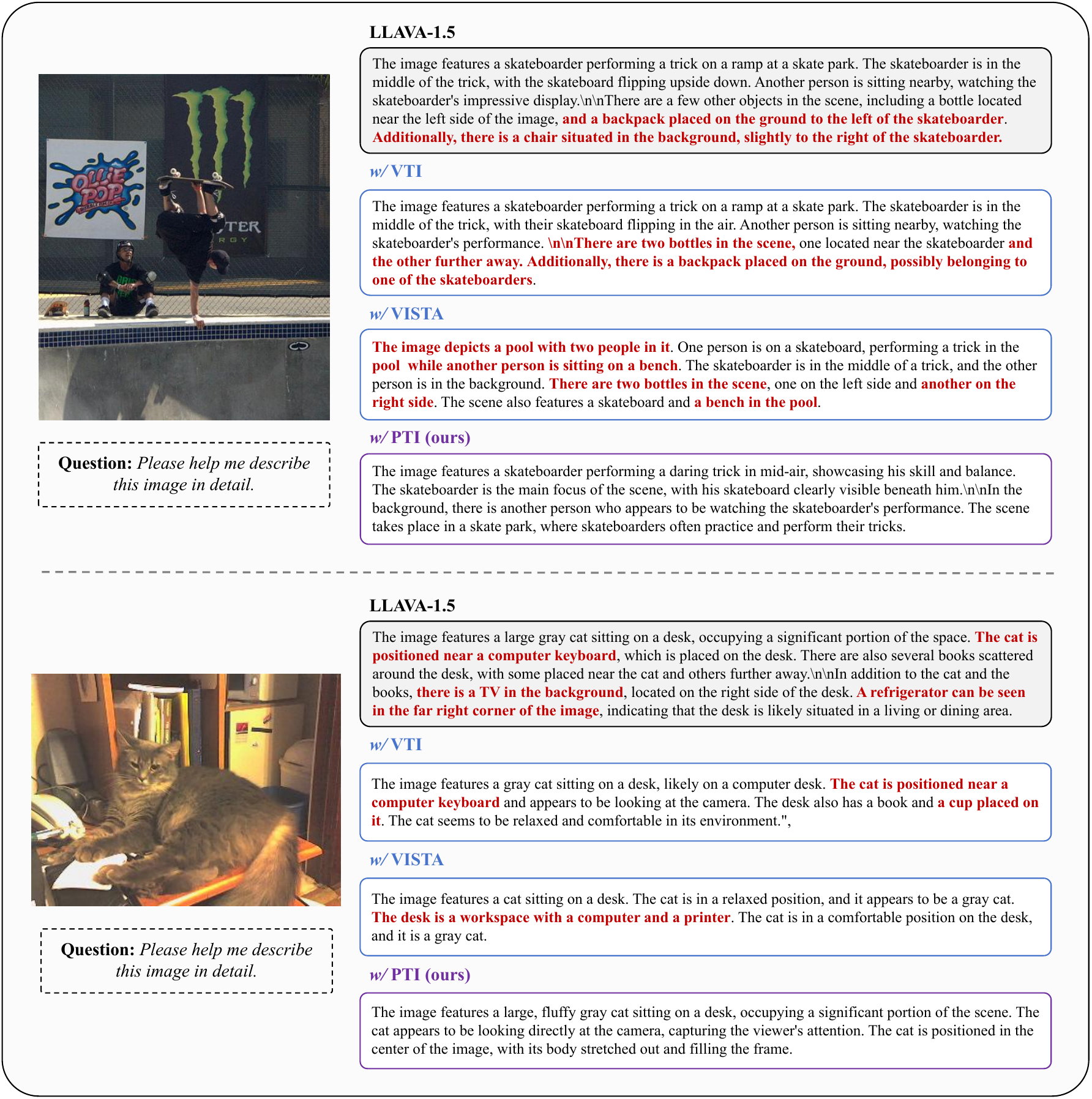} 
    \end{minipage}
  
    \caption{Qualitative examples of LLAVA-1.5. Hallucinated contents are marked in \textcolor{mydarkred}{\textbf{red}}.}
    \label{tables:llava_case} 
\end{figure*}

\begin{figure*}[htbp] 
    \centering

    \begin{minipage}[t]{\linewidth}
        \centering
        \includegraphics[width=\linewidth]{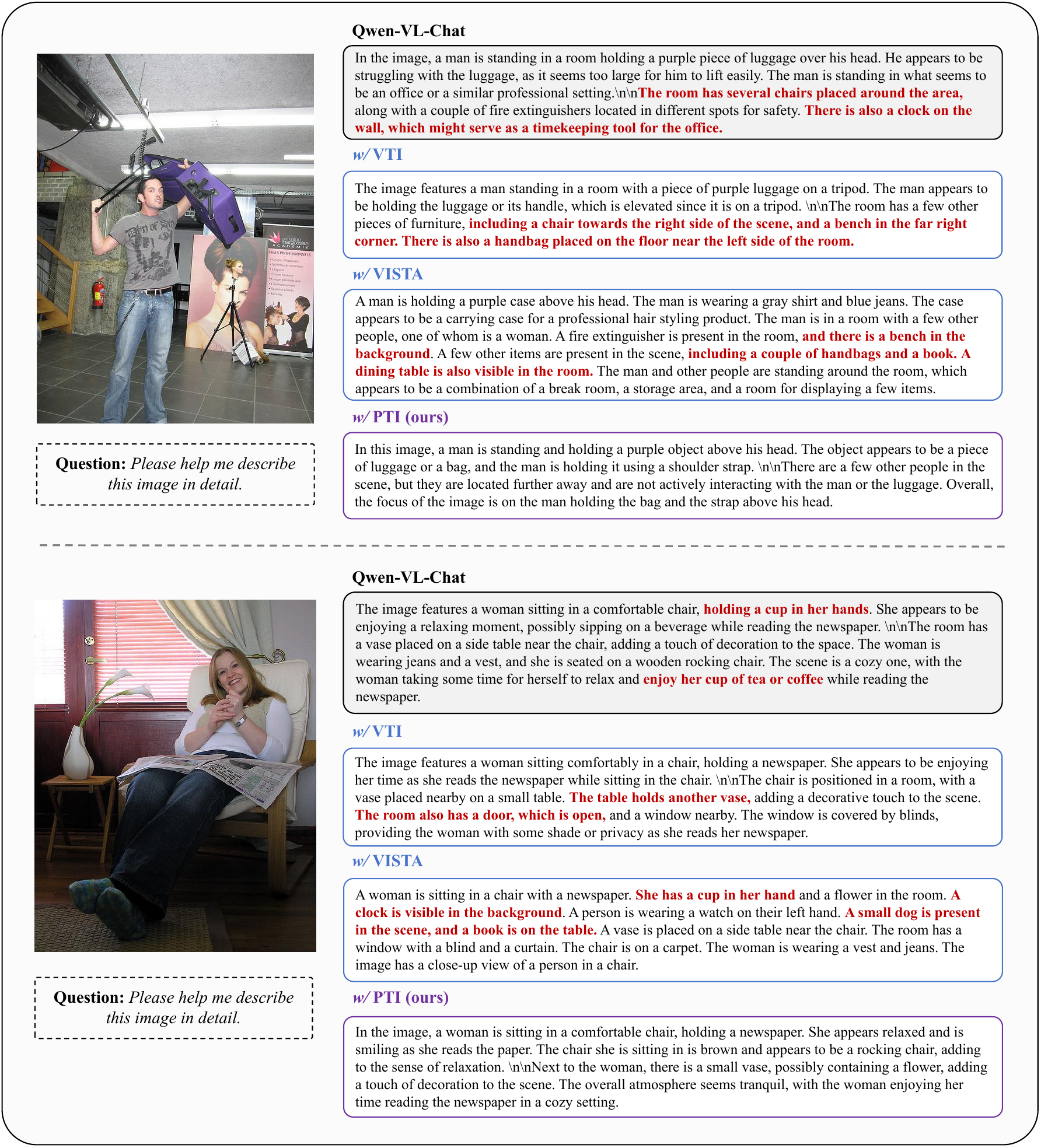} 
    \end{minipage}
  
    \caption{Qualitative examples of Qwen-VL-Chat. Hallucinated contents are marked in \textcolor{mydarkred}{\textbf{red}}.}
    \label{tables:qwen_case} 
\end{figure*}

\begin{figure*}[htbp] 
    \centering

    \begin{minipage}[t]{\linewidth}
        \centering
        \includegraphics[width=\linewidth]{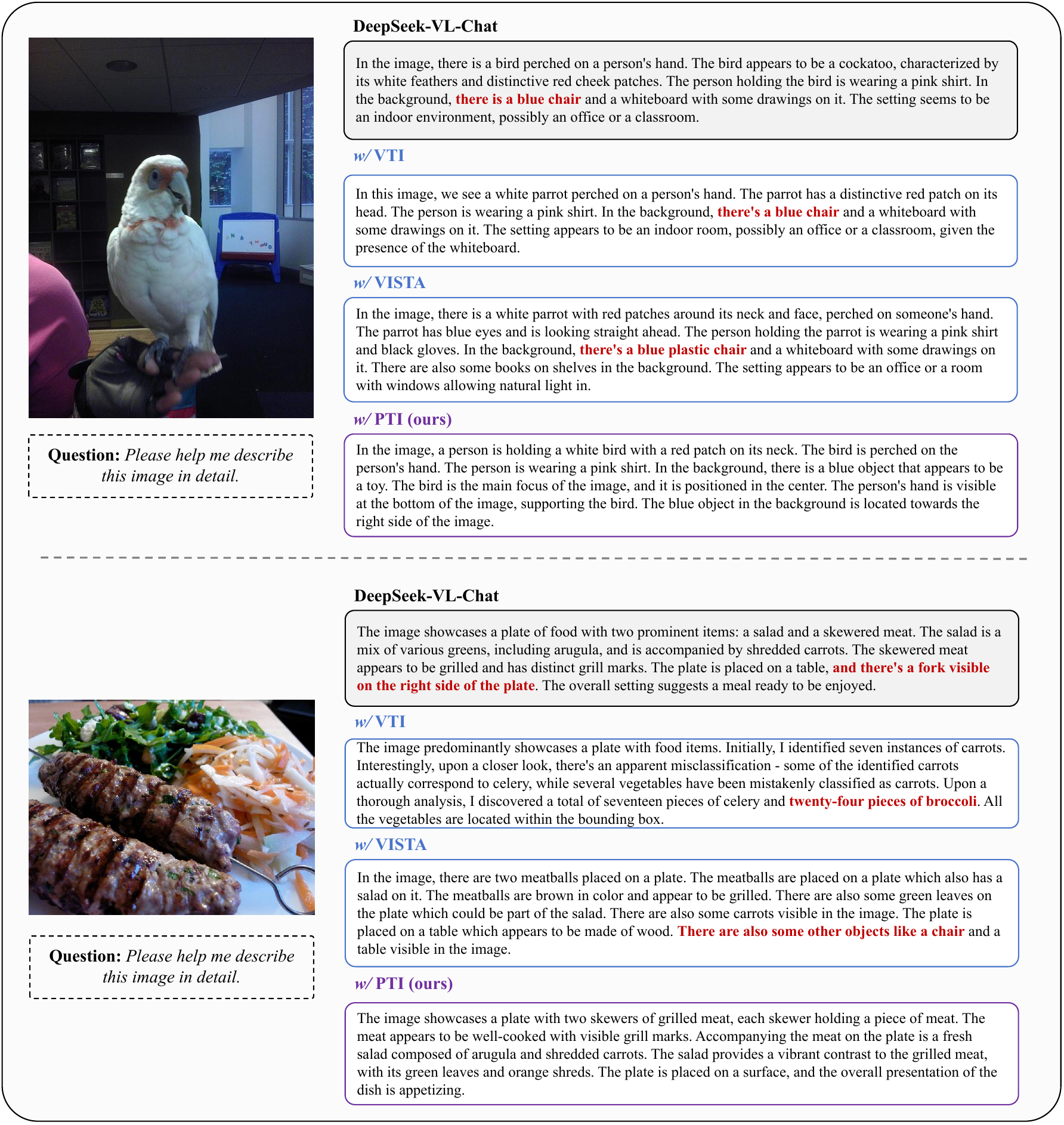} 
    \end{minipage}
  
    \caption{Qualitative examples of DeepSeek-VL-Chat. Hallucinated contents are marked in \textcolor{mydarkred}{\textbf{red}}.}
    \label{tables:deepseek_case} 
\end{figure*}

We further extend our qualitative evaluation across all three distinct architectures to demonstrate PTI’s effectiveness in reducing hallucination. 
Figures \ref{tables:llava_case}-\ref{tables:deepseek_case} present comparative examples between vanilla models, DTI methods (\textit{i.e.}, VTI \cite{liu2024reducing} and VISTA \cite{li2025hidden}), and our PTI for LLAVA-1.5, Qwen-VL-Chat, and DeepSeek-VL-Chat, respectively. 
As evident across these scenarios, while vanilla models and DTI methods frequently suffer from severe object hallucinations and context misinterpretation, PTI effectively suppresses the generation of non-existent entities and erroneous attributes. 
These qualitative examples demonstrate the superior ability of PTI to mitigate hallucination in large vision-language models by purifying initial representations before decoding.




\end{document}